\documentclass{article}

\usepackage[table]{xcolor}
\usepackage{subcaption}

\usepackage{microtype}
\usepackage{booktabs} %

\usepackage[accepted]{icml2024}

\usepackage{amsmath}
\usepackage{amssymb}
\usepackage{mathtools}
\usepackage{amsthm}

\theoremstyle{plain}
\newtheorem{theorem}{Theorem}[section]

\theoremstyle{definition}
\newtheorem{definition}[theorem]{Definition}

\theoremstyle{remark}

\usepackage[textsize=tiny]{todonotes}
\usepackage{algorithm}
\usepackage{multirow}
\usepackage{enumitem}
\usepackage{placeins}
\newcommand{\copyed}[1]{\textcolor{black}{#1}} %

\definecolor{CColor}{rgb}{0.01,0.31,0.59}
\definecolor{GGray}{rgb}{0.80,0.90,1}
\definecolor{Shady}{rgb}{0.9,0.9,0.9}
\definecolor{kaistblue}{RGB}{20,135,200}
\definecolor{kaistdarkblue}{RGB}{0,65,145}
\definecolor{urbanablue}{RGB}{19,41,75}
\definecolor{urbanaorange}{RGB}{232,74,39}
\definecolor{drp}{rgb}{0.53,0.15,0.34}
\usepackage[plainpages=false,pdfpagelabels,colorlinks=true,linkcolor=CColor,citecolor=CColor,urlcolor=CColor]{hyperref}

\usepackage{titletoc}
\usepackage{tocloft}
\DeclareMathOperator*{\argmin}{arg\,min}

\icmltitlerunning{Merging Multi-Task Models via Weight-Ensembling Mixture of Experts}

\begin{document}

\twocolumn[
  \icmltitle{
    Merging Multi-Task Models via Weight-Ensembling Mixture of Experts}

  \icmlsetsymbol{equal}{*}

  \begin{icmlauthorlist}
    \icmlauthor{Anke Tang}{whu,luojia,equal}
    \icmlauthor{Li Shen}{sysu,jd,equal}
    \icmlauthor{Yong Luo}{whu,luojia}
    \icmlauthor{Nan Yin}{mbzuai}
    \icmlauthor{Lefei Zhang}{whu,luojia}
    \icmlauthor{Dacheng Tao}{ntu}
  \end{icmlauthorlist}

  \icmlaffiliation{whu}{National Engineering Research Center for Multimedia Software, School of Computer Science, Wuhan University, China}
  \icmlaffiliation{luojia}{Hubei Luojia Laboratory, Wuhan, China}
  \icmlaffiliation{sysu}{Sun Yat-sen University, Shenzhen, China}
   \icmlaffiliation{jd}{JD Explore Academy, China}
  \icmlaffiliation{mbzuai}{Mohamed bin Zayed University of Artificial Intelligence, United Arab Emirates}
  \icmlaffiliation{ntu}{Nanyang Technological University, Singapore}

  \icmlcorrespondingauthor{Yong Luo}{luoyong@whu.edu.cn}

  \icmlkeywords{Machine Learning, ICML}

  \vskip 0.3in
]

\printAffiliationsAndNotice{\icmlEqualContribution} %

\begin{abstract}
  Merging various task-specific Transformer-based vision models trained on different tasks into a unified model can execute all the tasks concurrently.
  Previous methods, exemplified by task arithmetic, have proven to be both effective and scalable.
  Existing methods have primarily focused on seeking a static optimal solution within the original model parameter space. A notable challenge is mitigating the interference between parameters of different models, which can substantially deteriorate performance.
  In this paper, we propose to merge most of the parameters while upscaling the MLP of the Transformer layers to a weight-ensembling mixture of experts (MoE) module, which can dynamically integrate shared and task-specific knowledge based on the input, thereby providing a more flexible solution that can adapt to the specific needs of each instance.
  Our key insight is that by identifying and separating shared knowledge and task-specific knowledge, and then dynamically integrating them, we can mitigate the parameter interference problem to a great extent.
  We conduct the conventional multi-task model merging experiments and evaluate the generalization and robustness of our method.
  The results demonstrate the effectiveness and provide a comprehensive understanding of our method.
\end{abstract}

\begin{figure*}[t]
  \begin{center}
    \centerline{\includegraphics[width=\linewidth]{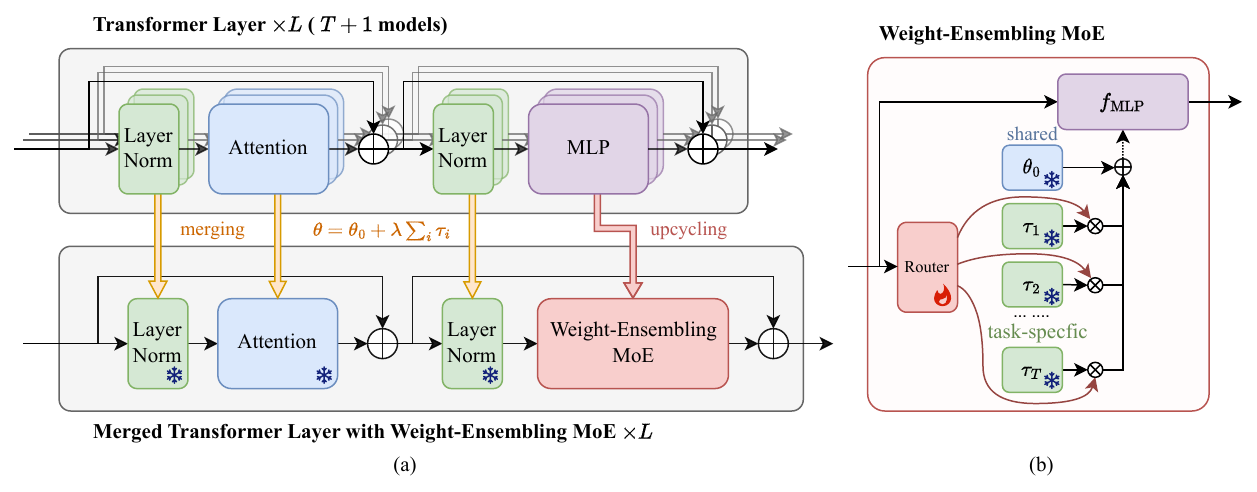}}
    \vskip -0.1in
    \caption{
      (a) \textbf{Framework overview}. This figure shows the overall framework of our proposed method to merge the pre-trained model and fine-tuned task-specific models. We merge weights in the Transformer Layers except for the MLPs. For the MLPs, we upcycle them into weight-assembling MoE modules.
      (b) \textbf{Wieght-Ensembling Mixture of Experts (MoE) Module}. Here we outline the detailed structure of the Weight-Ensembling MoE module, composed of the router, pre-trained MLP weights, and a collection of task vectors. Collaboration between shared weights and task vectors is employed to create input-conditioned weights dynamically. In this way, we separate shared information and task-specific knowledge, which are then combined based on input in time.
    }
    \label{fig:overview}
  \end{center}
  \vskip -0.2in
\end{figure*}

\section{Introduction}
\label{section:introduction}

The swift advancement of deep learning has fostered a shift towards fine-tuning large pre-trained models for downstream tasks, rather than training them from scratch. Having been initially trained on large-scale datasets, pre-trained models have outstanding common sense and are adept at recognizing and processing diverse data patterns~\citep{radfordLanguageModelsAre2019,heMaskedAutoencodersAre2021,DBLP:journals/ijautcomp/WangCQGWWTG23}. These models can then be fine-tuned on downstream tasks to acquire task-specific knowledge~\citep{chungScalingInstructionFinetunedLanguage2022,zhengLearnModelFineTuning2023,DBLP:journals/ijautcomp/CaoLHS24}. In this context, merging multiple task-specific models into a single unified model has emerged as an effective and scalable strategy for knowledge transfer and multi-task learning~\citep{liDeepModelFusion2023,DBLP:journals/ijautcomp/LinGGZZL23}.

There are several state-of-the-art algorithms for merging models. A prominent example in this field is task arithmetic~\citep{ilharcoEditingModelsTask2023}, which interpolates the parameters of models linearly.
These methods excel in extracting knowledge from various models and synthesizing a unified model cheaply. 
Such approaches present a promising solution for constructing robust models~\citep{izmailovAveragingWeightsLeads2019,wortsmanModelSoupsAveraging2022}, especially in the scenarios where the training data is decentralized, limited, or inaccessible due to privacy constraints~\citep{tangImprovingHeterogeneousModel2023a,jinDatalessKnowledgeFusion2023}. 

Existing methods predominantly aim to find a static multi-task optimal solution within the original parameter space. 
These methods do not introduce any new parameters, thus maintaining the original inference cost. This approach, however, imposes limitations on adaptability to the unique requirements of each instance, as the task-specific optimal solution varies.
Another significant challenge of merging multi-task models is mitigating the interference between parameters of different models, which can substantially deteriorate the average performance~\citep{yadavResolvingInterferenceWhen2023,yuLanguageModelsAre2023,tangConcreteSubspaceLearning2023}. 
Existing methods, while effective in some scenarios, may not be flexible enough to handle the dynamic nature of multi-task learning, where the optimal solution can vary depending on the input.

To address these challenges, we propose a novel approach to merge vision Transformers (ViTs). 
Our method merges most of the parameters while upscaling the multilayer perceptron (MLP) of the Transformer layers to a weight-ensembling Mixture of Experts (MoE) module. This module can dynamically integrate shared and task-specific knowledge based on the input sample, thereby providing a more flexible solution that can adapt to the specific needs of each instance.

Our primary realization is that the issue of parameter interference can be significantly alleviated by identifying and separating shared and task-specific knowledge and then dynamically integrating them. 
So we can leverage the shared knowledge that is beneficial across all tasks, while also taking into account the unique requirements. 
By dynamically combining these two types of knowledge based on the specific input data, we can create a more flexible and adaptable model that can effectively handle a wide range of tasks. 

We validate our method through conventional multi-task model merging experiments and evaluate its generalization and robustness. The results demonstrate the effectiveness of our method and provide a comprehensive understanding.

To summarize, our contributions are as follows:
\begin{itemize}
  \item We propose a novel method to merge Transformer-based models. Our method is effective in transferring knowledge from various task-specific fine-tuned models and constructing a unified multi-task model.
  \item We design a novel Weight-Ensembling MoE (WEMoE) module, which can dynamically integrate shared and task-specific knowledge based on the input sample.
  \item We conduct extensive experiments, and the results demonstrate the effectiveness of our method and provide a comprehensive understanding of our method.
\end{itemize}

\section{Revisiting Model Merge for MTL}

In this section, we first introduce the problem setting and notations. Then we revisit the model merge methods for multi-task learning and discuss their limitations.

\subsection{Problem Formulation}
\label{subsection:problem_formulation}

We begin with a large pre-trained neural network, which is parameterized by $\theta_0 \in \mathbb{R}^{|\theta|}$, which is adaptable to a wide range of downstream tasks through fine-tuning.
Given a set of $n$ downstream tasks, denoted as $\mathcal{S} = \{s_i\}_{i=1}^n$, we fine-tune the pre-trained model $f$ individually for each task $s_i$, resulting in a series of fine-tuned models, each characterized by its unique parameters $\theta_i$.
Our goal is to merge the pre-trained model $f$ and the fine-tuned models $\{f_i\}_{i=1}^n$ into a single model $f_{\text{merged}}$ that can handle all the tasks in $\mathcal{S}$.

Before providing an overview of our framework and delving into the details, we first need to introduce the concept of task vector, which is a critical element of our method.
\begin{definition}[Task Vector~\citep{ilharcoEditingModelsTask2023}]
  Task vector $\tau_i$ is defined as the difference between the parameters of the  fine-tuned model and the pre-trained model, i.e.,
  \begin{equation}
    \label{eq:task_vector}
    \tau_i = \theta_i - \theta_0.
  \end{equation}
\end{definition}

\subsection{Revisiting Model Merge}
\label{subsection:revisiting_model_merge}

From a multi-objective optimization perspective, the solution space of the optimal merged models is the Pareto front of the downstream tasks in $\mathcal{S}$. In other words, an optimal merged model should be able to achieve the best performance for all tasks in $\mathcal{S}$ simultaneously.
It is challenging to find the optimal merged model, as the solution space is large and complex. What's more, we do not have access to the training data of the downstream tasks but only the pre-trained model and the fine-tuned models, which makes it impossible to train a model from scratch as in the traditional multi-task learning setting.
We thus need to find a way to transfer the knowledge from the pre-trained model and the fine-tuned models into a unified merged model.

\textbf{Limitation of static solutions.}
Current merging methodologies predominantly aim to find a static solution within the original model parameter space. This approach, however, restricts their adaptability to the unique requirements of each instance, as the optimal solution can depend on the input.
Finding a Pareto optimal solution in multi-objective optimization can lead to suboptimal performance on individual objectives. As the concept of Pareto optimality is rooted in the idea that a solution is Pareto optimal if there is no static solution that simultaneously improves one objective without degrading at least one other objective. For example, we can not minimize the loss of all tasks better than the global optimal of the joint loss function $\argmin_\theta \sum_{i=1}^n \mathcal{L}_i(\theta)$, where $\mathcal{L}_i$ is the loss function of single downstream task $s_i$.

\begin{figure}[t]
  \vskip 0.1in
  \begin{center}
    \includegraphics[width=1.05\linewidth]{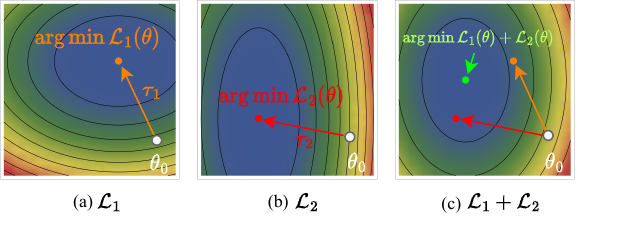}
    \vskip -0.15in
    \caption{
      Illustration of the loss landscapes of $s_1$, $s_2$, and $s_1 \cup s_2$.
      There is no static solution $\theta'$ that simultaneously minimizes the loss of both tasks better than $\argmin_\theta \mathcal{L}_1(\theta) + \mathcal{L}_2(\theta)$.
    }
    \label{fig:loss_landscapes}
  \end{center}
  \vskip -0.2in
\end{figure}

This concept is visually represented in Figure~\ref{fig:loss_landscapes}, where the loss landscapes of $s_1$, $s_2$, and $s_1 \cup s_2$ are shown.
No static solution $\theta'$ that simultaneously satisfies $\mathcal{L}_1(\theta') < \mathcal{L}_1(\theta^*)$ and $\mathcal{L}_2(\theta') < \mathcal{L}_2(\theta^*)$, where $\theta^* = \argmin_\theta \mathcal{L}_1(\theta) + \mathcal{L}_2(\theta)$.
However, $\theta^*$ is a suboptimal solution for both tasks individually, as $\mathcal{L}_1(\theta^*) > \min_\theta \mathcal{L}_1(\theta)$ and $\mathcal{L}_2(\theta^*) > \min_\theta \mathcal{L}_2(\theta)$.
Visualization of loss landscapes in Figure~\ref{fig:loss_landscapes_examples}, Appendix~\ref{appendix:loss}.

It's important to note that there can be instances where two tasks are sufficiently related to positively impact each other, especially in specific multi-task learning and auxiliary-task learning scenarios. However, this heavily depends on the task set and domain.
In our study, we primarily focus on situations where negative transfer is a prevalent concern. In our case, tasks from various domains are not closely related, which creates a significant domain gap.

\textbf{Knowledge separation.}
Common approaches to knowledge separation include knowledge distillation~\citep{hintonDistillingKnowledgeNeural2015}, pruning~\citep{franklePruningNeuralNetworks2021}, and feature extraction~\citep{yosinskiUnderstandingNeuralNetworks2015}.
It often requires heavy computation to separate or extract the knowledge from deep neural networks.
However, knowledge separation is a crucial step for model merging, as it allows us to leverage the shared knowledge that is beneficial across all tasks, while also taking into account the unique requirements of each task.
Here, we aim to separate the shared knowledge and task-specific knowledge computationally cheaply and data-freely.

Fortunately, as evident from Eq.(\ref{eq:task_vector}), the task vector represents the modifications applied to the pre-trained model to optimize it for a specific task $s_i$. Therefore, it can be interpreted as encapsulating the knowledge specific to that task naturally.
On the other hand, the pre-trained model, characterized by its parameters $\theta_0$, encapsulates the shared knowledge that is relevant across all tasks in the set $\mathcal{S}$. This shared knowledge is derived from the large dataset on which the model was initially trained, and it forms the foundation upon which task-specific adjustments are made.

\begin{figure}[t]
  \vskip 0.1in
  \begin{center}
    \begin{subfigure}{0.46\linewidth}
      \centering
      \includegraphics[width=\linewidth]{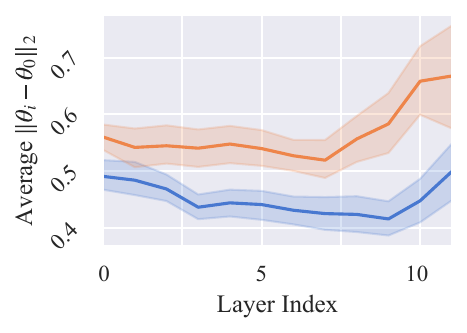}
      \caption{$L_2$ length of task vectors}
    \end{subfigure}
    \begin{subfigure}{0.49\linewidth}
      \centering
      \includegraphics[width=\linewidth]{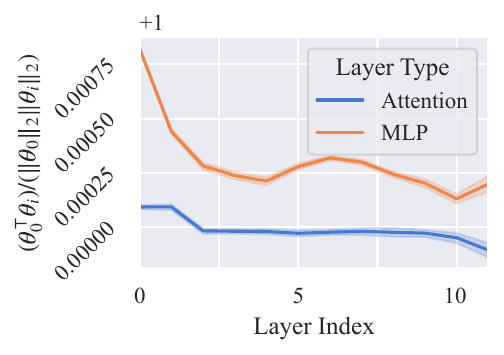}
      \caption{cosine similarity}
    \end{subfigure}
    \vskip -0.1in
    \caption{
      The distance between the parameters of the pre-trained model and the fine-tuned models (CLIP-ViT-B/32 on eight tasks).
    }
    \label{fig:distance_between_parameters}
  \end{center}
  \vskip -0.2in
\end{figure}

\textbf{Similarity in parameter space.}
Another consideration we must address is the choice of the model segment to undergo knowledge separation. Applying knowledge separation across the entire model might yield a model overly tailored to a specific task or one that is computationally intensive during knowledge reintegration.
Conversely, if we limit knowledge separation to a small portion of the model, we may not be able to separate the shared knowledge and task-specific knowledge effectively, leading to a suboptimal merged model.
Striking a balance between computational cost and performance necessitates identifying the most valuable segment for knowledge separation.

Empirically, we found that for Transformer models, the weights of the Attention modules in the fine-tuned models have a higher similarity to those in the pre-trained model than MLP modules, consistent with findings in \cite{lawsonMergingDecisionTransformers2023}.
In Figure~\ref{fig:distance_between_parameters}, we illustrate the distance between the parameters of the pre-trained model and the fine-tuned models (CLIP-ViT-B/32 on eight tasks) using $L_2$ norm and cosine similarity.
As shown in Figure~\ref{fig:distance_between_parameters}(a), the $L_2$ norm indicates MLPs have higher similarity. However, in Figure~\ref{fig:distance_between_parameters}(b), Attention modules have higher cosine similarity, suggesting a closer relationship.
The discrepancy arises because $L_2$ norm assesses magnitude changes while cosine similarity measures directional alignment.
The magnitude of parameters in a neural network plays a crucial role in determining the model's performance and generalization ability. Parameters with large magnitudes can lead to overfitting, where the model learns to fit the noise in the training data, resulting in poor performance on unseen data. On the other hand, parameters with small magnitudes can lead to underfitting, where the model fails to capture the underlying patterns in the data, resulting in suboptimal performance.
The $L_2$ distance directly measures the differences in parameter magnitudes, providing a clear indication of how much the model's performance might change.
To this end, we use the $L_2$ norm to measure the functional similarity between the parameters of fine-tuned and pre-trained models.

\section{Methodology}
\label{section:methodology}

In this section, we provide an overview of our proposed framework before delving into the detailed workings of the Weight-Ensembling Mixture of Experts (MoE) module. Finally, we discuss the test-time adaptation training process.

\subsection{Framework Overview}
\label{subsection:framework_overview}

From the previous observations and discussion, our key insight is that we can separate the shared information and task-specific knowledge and combine them dynamically in parameter space based on inputs.
Based on this insight, we propose a novel framework to merge the pre-trained model and the fine-tuned models, as shown in Figure~\ref{fig:overview}(a).

So far we have identified the shared information and task-specific knowledge. The next step is to combine them in a way that allows the merged model to handle all the tasks in $\mathcal{S}$. The ability to separate and leverage both knowledge is a key strength of our proposed framework.
To achieve this, we propose to dynamically integrate the shared information and task-specific knowledge based on the input samples, rather than seeking a static solution within the original parameter space. This dynamic adjustment allows the model to better adapt to the nuances of each task, thereby improving its performance across source tasks.

To this end, we propose to upcycle the MLPs in the pre-trained model and the fine-tuned models into Weight-Ensembling MoE modules, which are designed to dynamically select task-specific knowledge and combine it with the shared information based on the input samples. We will discuss the details of the Weight-Ensembling MoE module in Section \ref{subsection:weight_ensembling_moe_module}.
For the remaining parts of the model, we utilize the Task Arithmetic method~\citep{ilharcoEditingModelsTask2023} to merge the weights. Task Arithmetic is a simple yet effective and scalable method for merging models, it operates in the parameter space element-wise without modification to the model structure.
The merged model has parameters $\theta = \theta_0 + \lambda \sum_{i=1}^n \tau_i$, where $\lambda$ is a hyperparameter that controls the contribution of the task vectors to the model.

\subsection{Weight-Ensembling MoE Module}
\label{subsection:weight_ensembling_moe_module}

In this subsection, we delve into the specifics of the Weight-Ensembling Mixture of Experts (MoE) module.
As depicted in Figure \ref{fig:overview} (b), the Weight-Ensembling MoE module is composed of three main components: the router, the pre-trained MLP weights, and a collection of task vectors.

The router $r: \mathbb{R}^d \rightarrow \mathbb{R}^T$, which is essentially a simple MLP, processes the input and subsequently generates a routing weights. These routing weights are used to determine how the knowledge from different tasks is combined.
The pre-trained MLP weights $\theta_0^{\text{MLP}}$ are the weights of the MLPs from the pre-trained model. These weights are crucial as they have been trained to recognize and process a wide range of data patterns, and each finetuned model encompasses this shared information.
The task vectors $\{\tau_i^{\text{MLP}}| \theta_i^{\text{MLP}} - \theta_0^{\text{MLP}}\}_{i=1}^T$, on the other hand, represent the differences between the MLPs that have been fine-tuned for specific tasks and the pre-trained ones, thus capture the unique adjustments made to the MLPs to optimize them for specific tasks and contain the task-specific knowledge.

To summarize, our proposed Weight-Ensembling MoE module is designed to segregate shared information and task-specific knowledge. This separation allows the module to handle a wide range of tasks without compromising the general applicability of the shared information. The shared information and task-specific knowledge are then dynamically combined based on the input samples.

The mathematical representation of the weight-ensembling MoE module can be expressed as follows:
\begin{align}
  \label{eq:routing_weights}
  \left\{
  \begin{array}{l}
    w = \mathop{mean}(r(\mathbf{h}_{1:N}^{\text{in}})) \in \mathbb{R}^{T\times 1}, \\
    \theta^{\text{MLP}} = \theta_0^{\text{MLP}} + \mathbf{D}_{\tau} w,             \\
    \mathbf{h}_{1:N}^{\text{out}} = f_{\text{MLP}}(\mathbf{h}_{1:N}^{\text{in}}; \theta^{\text{MLP}}).
  \end{array}
  \right.
\end{align}
Where $\mathbf{h}_{1:N}^{\text{in}}$ is the input sequence of tokens, $\mathbf{h}_{1:N}^{\text{out}}$ is the output sequence of tokens, $f_{\text{MLP}}$ is the MLP function and $\mathbf{D}_{\tau} \in \mathbb{R}^{|\theta^{\text{MLP}}|\times T}$ is the dictionary matrix of task vectors, which is made up of fixed task vectors. The $\mathop{mean}$ function averages the routing weights across the tokens in the input sequence.
From the perspective of dictionary learning, the Weight-Ensembling MoE module can be viewed as a dictionary look-up operation, where the routing weights are used to select the task vectors from the dictionary matrix $\mathbf{D}_{\tau}$, which are then added to the pre-trained MLP weights $\theta_0^{\text{MLP}}$ to create the input-conditioned weights $\theta^{\text{MLP}}$.

\textbf{The structure and initialization of the router.}
In our weight-ensembling MoE module, the router plays a crucial role.
It is responsible for directing the input data to the appropriate combination of expert task vectors.
In our research, we employed a straightforward n-layer MLP as the router.
During our experiments, we explored two configurations for the MLP: one with no hidden layers ($l=0$) and another with two hidden layers ($l=2$). The mathematical representation of these configurations is as follows:
\begin{equation}
  \label{eq:leq2}
  r(h) = \mathbf{W}_2 \mathop{ReLU}(\mathbf{W}_1 h + \mathbf{b}_1) + \mathbf{b}_2, \quad l=2,
\end{equation}
and
\begin{equation}
  \label{eq:leq0}
  r(h) = \mathbf{b}_0, \quad l=0.
\end{equation}
Where $\mathbf{W}$ and $\mathbf{b}$ are the weight and bias, respectively, and $\mathop{ReLU}$ is the rectified linear unit activation function.
Unless otherwise specified, we utilize a structure with $l=2$.
In Appendix~\ref{appendix:ablations_of_router_depth}, we provide an in-depth analysis of the router's structure and its impact on the model's performance.

Initializing the router is crucial to setting up the MoE module, as it can greatly influence its performance. To ensure that the initial routing weights are approximately equal to $\lambda$, we initialize $\mathbf{W}_1$ and $\mathbf{W}_2$ by sampling from a Gaussian distribution with a mean of $0$ and a variance of $0.01$. We then set $\mathbf{b}_1$ and $\mathbf{b}_2$ to zeros and $\lambda$, respectively. In the case where $l=0$, we assign $\lambda$ to $\mathbf{b}_0$.
In fact, for a router with $l=0$, we can consider it as a partial implementation of AdaMerging~\citep{yangAdaMergingAdaptiveModel2023}, where only the MLPs undergo adaptive merging weights search.

\subsection{Test-Time Adaptation Training}
\label{subsection:test_time_adaptation_training}

Once the router has been initialized, the next step is to fine-tune its parameters. However, we have no access to the training data of the downstream tasks. To achieve this, we employ test-time adaptation training techniques, which are widely used in the field of semi-supervised learning~\citep{mounsavengBagTricksFully2023,liangComprehensiveSurveyTestTime2023}.
Test-time adaptation is a powerful technique that allows the model to adjust its parameters based on the unlabeled test data, thereby improving the performance of the merged model.

In our research, we focus on classification tasks and aim to minimize the multi-task entropy loss of the merged model on the unlabeled test data. Entropy loss is a measure of the uncertainty of the predictions. It is defined as follows:
\begin{align}
  \label{eq:entropy_loss}
  \mathcal{L}_{\text{entropy}}
   & = \mathbb{E}_{x\sim \mathcal{D}_{\text{test}}}[-p(\hat{y}|x) \log p(\hat{y}|x)]           \\
   & \approx -\frac{1}{|D|}\sum_{i=1}^{|D|} \sum_{c=1}^C p(\hat{y}_c|x_i) \log p(\hat{y}_c|x).
\end{align}
This is the empirically estimated entropy loss, where $\mathcal{D}_{\text{test}}$ represents the test data distribution, $p(\hat{y}|x)$ is the predicted posterior probability distribution, $C$ is the total number of classes, and $|D|$ is the number of samples in the test dataset.
Minimizing it encourages the model to make predictions with higher confidence. This can lead to improved performance on the test data, as the model is more likely to make correct predictions when it is confident in its decisions.

\section{Experiments}
\label{section:experiments}

In this section, we conduct experiments to evaluate the performance of our proposed method, including the conventional multi-task model merging experiments, and ablation studies to evaluate the generalization ability and robustness.
The code is available at \url{https://github.com/tanganke/weight-ensembling_MoE}.

\begin{table}[t]
  \caption{Parameter count of the up-scaled models from eight tasks.}
  \label{table:parameter_counts}
  \vskip 0.15in
  \begin{center}
    \begin{small}
      \begin{sc}
        \begin{tabular}{lccc}
          \toprule
          \textbf{Model} & \textbf{Trainable} & \textbf{Total} & \textbf{Ratio} \\
          \midrule
          \multicolumn{4}{c}{$l=0$}                                             \\
          CLIP-ViT-B/32  & 96                 & 566.80M        & 0.00\%         \\
          CLIP-VIT-B/16  & 96                 & 565.15M        & 0.00\%         \\
          CLIP-VIT-L/14  & 192                & 1.95B          & 0.00\%         \\
          \midrule
          \multicolumn{4}{c}{$l=2$}                                             \\
          CLIP-ViT-B/32  & 7.16M              & 573.96M        & 1.25\%         \\
          CLIP-VIT-B/16  & 7.16M              & 572.31M        & 1.25\%         \\
          CLIP-VIT-L/14  & 25.39M             & 1.98B          & 1.28\%         \\
          \bottomrule
        \end{tabular}
      \end{sc}
    \end{small}
  \end{center}
\end{table}
\begin{figure}[t]
  \begin{center}
    \begin{subfigure}{0.49\linewidth}
      \centering
      \includegraphics[width=\linewidth]{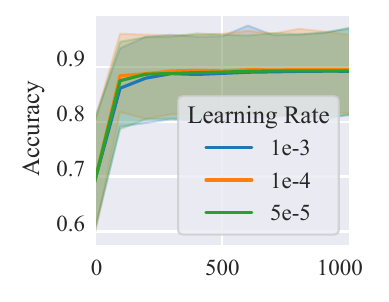}
      \caption{Learning rate comparison.}
      \label{fig:b32_fusion}
    \end{subfigure}
    \begin{subfigure}{0.49\linewidth}
      \centering
      \includegraphics[width=\linewidth]{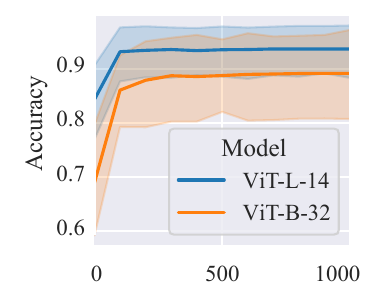}
      \caption{Model comparison.}
      \label{fig:clip_b32_and_l14_fusion}
    \end{subfigure}
    \caption{The performance of the merged models with a varying number of steps.
      (a) CLIP-ViT-B/32 model with different learning rate.
      (b) Comparison of CLIP-ViT-B/32 and CLIP-ViT-L/14.
    }
    \label{fig:wemoe_fusion}
  \end{center}
  \vskip -0.3in
\end{figure}

\begin{table*}[t]
  \caption{Multi-task performance when merging CLIP-ViT-B/32 models on all eight tasks.}
  \label{table:multi-task_performance_clip-vit-b-32}
  \vskip 0.1in
  \begin{center}
    \setlength{\tabcolsep}{1.5mm}
    \begin{small}
      \begin{sc}
        \begin{tabular}{lccccccccc}
          \toprule
          \textbf{Method}       & \textbf{SUN397} & \textbf{Cars} & \textbf{RESISC45} & \textbf{EuroSAT} & \textbf{SVHN} & \textbf{GTSRB}         & \textbf{MNIST} & \textbf{DTD}    & \textbf{Avg.} \\
          \midrule
          Pre-trained           & 63.2            & 59.6          & 60.2              & 45.0             & 31.6          & 32.6                   & 48.3           & 44.4            & 48.1          \\
          Individual            & \textbf{75.3}   & \textbf{77.7} & \textbf{96.1}     & \textbf{99.9}    & \textbf{97.5} & 98.7                   & \textbf{99.7}  & \textbf{79.4}   & \textbf{90.5} \\
          Traditional MTL       & \copyed{73.9}   & \copyed{74.4} & \copyed{93.9}     & \copyed{98.2}    & \copyed{95.8} & \textbf{\copyed{98.9}} & \copyed{99.5}  & \copyed{77.9}   & \copyed{88.9} \\
          \midrule
          \multicolumn{10}{c}{\textit{Multi-Task Model Fusion Methods}}                                                                                                                              \\
          Weight Averaging      & 65.3            & 63.3          & 71.4              & 73.6             & 64.2          & 52.8                   & 87.5           & 50.1            & 66.0          \\
          Fisher Merging        & \copyed{68.6}   & \copyed{69.2} & \copyed{70.7}     & \copyed{66.4}    & \copyed{72.9} & \copyed{51.1}          & \copyed{87.9}  & \copyed{59.9}   & \copyed{68.3} \\
          RegMean               & \copyed{65.3}   & \copyed{63.5} & \copyed{75.6}     & \copyed{78.6}    & \copyed{78.1} & \copyed{67.4}          & \copyed{93.7}  & \copyed{52.0}   & \copyed{71.8} \\
          Task Arithmetic       & 55.3            & 54.9          & 66.7              & 77.4             & 80.2          & 69.7                   & 97.3           & 50.1            & 69.0          \\
          Ties-Merging          & {65.0}          & {64.3}        & 74.7              & 76.8             & 81.3          & 69.4                   & 96.5           & {54.3}          & 72.8          \\
          AdaMerging (task)     & 58.3            & 53.2          & 71.8              & 80.1             & 81.6          & 84.4                   & 93.4           & 42.7            & 70.7          \\
          AdaMerging++ (task)   & \copyed{60.8}   & \copyed{56.9} & \copyed{73.1}     & \copyed{83.4}    & \copyed{87.3} & \copyed{82.4}          & \copyed{95.7}  & {\copyed{50.1}} & \copyed{73.7} \\
          AdaMerging (layer)    & 64.2            & 69.5          & 82.4              & 92.5             & 86.5          & 93.7                   & 97.7           & 61.1            & 80.9          \\
          AdaMerging++ (layer)  & \copyed{66.6}   & \copyed{68.3} & \copyed{82.2}     & \copyed{94.2}    & \copyed{89.6} & \copyed{89.0}          & \copyed{98.3}  & \copyed{60.6}   & \copyed{81.1} \\
          \textbf{WEMoE (Ours)} & \textbf{74.1}   & \textbf{77.4} & \textbf{93.7}     & \textbf{99.1}    & \textbf{96.2} & \textbf{98.9}          & \textbf{99.6}  & \textbf{76.4}   & \textbf{89.4} \\
          \bottomrule
        \end{tabular}
      \end{sc}
    \end{small}
  \end{center}
  \vskip -0.1in
\end{table*}
\begin{table*}[t]
  \caption{Multi-task performance when merging CLIP-ViT-L/14 models on all eight tasks.}
  \label{table:multi-task_performance_clip-vit-l-14}
  \begin{center}
    \setlength{\tabcolsep}{1.5mm}
    \small
    \begin{sc}
      \begin{tabular}{lccccccccc}
        \toprule
        \textbf{Method}       & \textbf{SUN397} & \textbf{Cars} & \textbf{RESISC45} & \textbf{EuroSAT} & \textbf{SVHN} & \textbf{GTSRB}  & \textbf{MNIST} & \textbf{DTD}           & \textbf{Avg.} \\
        \midrule
        Pre-trained           & 68.2            & 77.9          & 71.3              & 61.3             & 58.4          & 50.6            & 76.4           & 55.4                   & 64.9          \\
        Individual            & \textbf{82.3}   & \textbf{92.4} & \textbf{97.4}     & \textbf{99.9}    & \textbf{98.1} & \textbf{99.2}   & \textbf{99.7}  & 84.1                   & \textbf{94.1} \\
        Traditional MTL       & \copyed{80.8}   & \copyed{90.6} & \copyed{96.3}     & \copyed{96.3}    & \copyed{97.6} & \copyed{99.1}   & \copyed{99.6}  & \textbf{\copyed{84.4}} & \copyed{93.5} \\
        \midrule
        \multicolumn{10}{c}{\textit{Multi-Task Model Fusion Methods}}                                                                                                                              \\
        Weight Averaging      & 72.1            & 81.6          & 82.6              & 91.4             & 78.2          & 70.6            & 97.0           & 62.8                   & 79.5          \\
        Fisher Merging        & \copyed{69.2}   & \copyed{88.6} & \copyed{87.5}     & \copyed{93.5}    & \copyed{80.6} & \copyed{74.8}   & \copyed{93.3}  & \copyed{70.0}          & \copyed{82.2} \\
        RegMean               & \copyed{73.3}   & \copyed{81.8} & \copyed{86.1}     & \copyed{97.0}    & \copyed{88.0} & \copyed{84.2}   & \copyed{98.5}  & \copyed{60.8}          & \copyed{83.7} \\
        Task Arithmetic       & 74.1            & 82.1          & 86.7              & 92.6             & 87.9          & 86.8            & 98.9           & 65.6                   & 84.4          \\
        Ties-Merging          & 75.0            & 84.5          & 88.0              & 94.3             & 85.7          & 82.1            & 98.7           & 67.7                   & 84.5          \\
        AdaMerging (layer)    & \copyed{79.0}   & \copyed{90.3} & \copyed{90.8}     & \copyed{96.2}    & \copyed{93.4} & {\copyed{98.0}} & \copyed{99.0}  & {\copyed{79.9}}        & \copyed{90.8} \\
        AdaMerging++ (layer)  & {\copyed{79.4}} & \copyed{90.3} & \copyed{91.6}     & {\copyed{97.4}}  & \copyed{93.4} & \copyed{97.5}   & \copyed{99.0}  & \copyed{79.2}          & \copyed{91.0} \\
        \textbf{WEMoE (Ours)} & \textbf{81.4}   & \textbf{92.6} & \textbf{95.4}     & \textbf{99.4}    & \textbf{97.7} & \textbf{99.3}   & \textbf{99.7}  & \textbf{83.7}          & \textbf{93.6} \\
        \bottomrule
      \end{tabular}
    \end{sc}
  \end{center}
  \vskip -0.1in
\end{table*}
\begin{table*}[t]
  \caption{Generalization results on two unseen tasks when merging ViT-B/32 models on six tasks.}
  \label{table:generalization_results_clip-vit-b-32}
  \vskip 0.1in
  \begin{center}
    \setlength{\tabcolsep}{1.5mm}
    \begin{small}
      \begin{sc}
        \begin{tabular}{l|ccccccc|ccc}
          \toprule
          \multirow{2}{*}{\textbf{Method}} & \multicolumn{7}{c|}{\textbf{Seen Tasks}} & \multicolumn{3}{c}{\textbf{Unseen Tasks}}                                                                                                                                          \\
                                           & SUN397                                   & Cars                                      & RESISC45      & DTD           & SVHN          & GTSRB         & \textbf{Avg.} & MNIST         & EuroSAT                & \textbf{Avg.} \\
          \midrule
          Task Arithmetic                  & 63.4                                     & 62.3                                      & 75.3          & 57.8          & 84.7          & 80.4          & 70.7          & 77.3          & 45.6                   & 61.4          \\
          Ties-Merging                     & 67.8                                     & 66.2                                      & 77.0          & 56.2          & 77.2          & 71.0          & 69.2          & 75.9          & 43.1                   & 59.5          \\
          AdaMerging                       & \copyed{65.2}                            & \copyed{65.9}                             & \copyed{88.5} & \copyed{61.1} & \copyed{92.2} & \copyed{91.5} & \copyed{77.4} & \copyed{84.0} & \textbf{\copyed{56.1}} & \copyed{70.0} \\
          AdaMerging++                     & \copyed{68.2}                            & \copyed{67.6}                             & \copyed{86.3} & 63.6          & \copyed{92.6} & \copyed{89.8} & \copyed{78.0} & \copyed{83.9} & \copyed{53.5}          & \copyed{68.7} \\
          \textbf{WEMoE (0-layer)}         & 63.8                                     & 61.2                                      & 78.4          & 56.4          & 89.1          & 92.2          & 73.5          & 78.6          & 49.7                   & 64.2          \\
          \textbf{WEMoE (2-layer)}         & \textbf{74.4}                            & \textbf{78.3}                             & \textbf{94.8} & \textbf{75.6} & \textbf{96.8} & \textbf{99.0} & \textbf{86.5} & \textbf{86.3} & 55.9                   & \textbf{71.1} \\
          \bottomrule
        \end{tabular}
      \end{sc}
    \end{small}
  \end{center}
  \vskip -0.1in
\end{table*}

\begin{table*}[t]
  \caption{Ablations of the test data distribution on ViT-B/32 (for all methods, $\lambda=0.3$).}
  \label{table:abalation_data_distribution_vit_b_32}
  \begin{center}
    \setlength{\tabcolsep}{1.5mm}
    \fontsize{8}{9}\selectfont
    \begin{sc}
      \begin{tabular}{l|ccccc|ccccc}
        \toprule
        \textbf{Method}          & \textbf{Cars}                                           & \textbf{EuroSAT}                                          & \textbf{RESISC45} & \textbf{GTSRB} & \textbf{Avg.} & \textbf{Cars} & \textbf{EuroSAT} & \textbf{RESISC45} & \textbf{GTSRB} & \textbf{Avg.} \\
        \midrule
                                 & \multicolumn{5}{c|}{{Clean Test Set}}                   & \multicolumn{5}{c}{{Corrupted Test Set (Motion Blur)}}                                                                                                                                                 \\
        Task Arithmetic          & 66.9                                                    & 94.7                                                      & 82.6              & 75.1           & 79.8          & 65.3          & 68.1             & 80.0              & 64.2           & 69.4          \\
        Ties-Merging             & 67.5                                                    & 83.7                                                      & 79.8              & 65.3           & 74.1          & 65.6          & 57.5             & 77.5              & 55.4           & 64.0          \\
        AdaMerging               & 73.7                                                    & 96.1                                                      & 85.8              & 96.3           & 88.0          & 71.2          & 74.6             & 82.7              & 94.1           & 80.6          \\
        \textbf{WEMoE (0-layer)} & 68.5                                                    & 94.6                                                      & 82.0              & 93.2           & 84.6          & 65.3          & 74.1             & 79.8              & 89.0           & 77.0          \\
        \textbf{WEMoE (2-layer)} & \textbf{78.8}                                           & \textbf{99.5}                                             & \textbf{95.4}     & \textbf{99.1}  & \textbf{93.2} & \textbf{78.1} & \textbf{79.7}    & \textbf{94.6}     & \textbf{97.8}  & \textbf{87.6} \\
        \midrule
                                 & \multicolumn{5}{c|}{Corrupted Test Set (Impluse Noise)} & \multicolumn{5}{c}{Corrupted Test Set (Gaussian Noise)}                                                                                                                                                \\
        Task Arithmetic          & 62.1                                                    & \textbf{49.1}                                             & 72.7              & 40.4           & 56.1          & 63.6          & \textbf{55.4}    & 75.9              & 49.4           & 61.1          \\
        Ties-Merging             & 62.1                                                    & 42.0                                                      & 70.4              & 34.9           & 52.3          & 64.1          & 50.3             & 74.5              & 39.8           & 57.2          \\
        AdaMerging               & 67.2                                                    & 30.8                                                      & 75.9              & 77.5           & 62.8          & 69.9          & 41.2             & 80.6              & 76.0           & 66.9          \\
        \textbf{WEMoE (0-layer)} & 63.8                                                    & 40.7                                                      & 74.3              & 66.7           & 61.4          & 66.1          & 45.5             & 78.3              & 66.9           & 64.2          \\
        \textbf{WEMoE (2-layer)} & \textbf{74.7}                                           & 11.6                                                      & \textbf{91.4}     & \textbf{91.7}  & \textbf{67.3} & \textbf{76.8} & 29.7             & \textbf{93.2}     & \textbf{78.2}  & \textbf{69.5} \\
        \midrule
                                 & \multicolumn{5}{c|}{Corrupted Test Set (Pixelate)}      & \multicolumn{5}{c}{Corrupted Test Set (Spatter)}                                                                                                                                                       \\
        Task Arithmetic          & 2.8                                                     & 41.5                                                      & 22.8              & 66.6           & 33.4          & 63.3          & \textbf{60.1}    & 73.9              & 54.3           & 62.9          \\
        Ties-Merging             & \textbf{4.1}                                            & 40.8                                                      & 20.6              & 57.1           & 30.6          & 64.4          & 50.8             & 71.4              & 44.3           & 57.8          \\
        AdaMerging               & 2.5                                                     & \textbf{53.8}                                             & 22.4              & 90.6           & \textbf{42.3} & 69.9          & 43.6             & 75.4              & 89.4           & 69.6          \\
        \textbf{WEMoE (0-layer)} & 2.1                                                     & 52.7                                                      & \textbf{23.4}     & 84.5           & 40.7          & 65.8          & 49.3             & 72.9              & 83.3           & 67.8          \\
        \textbf{WEMoE (2-layer)} & 0.4                                                     & 9.6                                                       & 2.2               & \textbf{97.0}  & 27.3          & \textbf{76.2} & 28.2             & \textbf{91.2}     & \textbf{96.0}  & \textbf{72.9} \\
        \midrule
                                 & \multicolumn{5}{c|}{Corrupted Test Set (Contrast)}      & \multicolumn{5}{c}{Corrupted Test Set (JPEG Compression)}                                                                                                                                              \\
        Task Arithmetic          & 66.0                                                    & 62.9                                                      & 75.9              & 70.6           & 68.9          & 66.5          & 72.3             & 82.2              & 60.0           & 70.3          \\
        Ties-Merging             & 66.8                                                    & 53.4                                                      & 75.9              & 61.5           & 64.4          & 67.5          & 60.4             & 80.0              & 50.1           & 64.5          \\
        AdaMerging               & 71.7                                                    & 69.8                                                      & 79.3              & 95.1           & {79.0}        & {70.9}        & 75.8             & 83.6              & 90.1           & 80.1          \\
        \textbf{WEMoE (0-layer)} & 67.3                                                    & 68.5                                                      & 74.8              & 91.4           & 75.5          & 66.4          & 75.3             & 81.4              & 83.1           & 76.5          \\
        \textbf{WEMoE (2-layer)} & \textbf{77.7}                                           & \textbf{77.4}                                             & \textbf{93.9}     & \textbf{98.5}  & \textbf{86.9} & \textbf{78.2} & \textbf{80.7}    & \textbf{95.1}     & \textbf{96.2}  & \textbf{87.6} \\
        \bottomrule
      \end{tabular}
    \end{sc}
  \end{center}
  \vskip -0.1in
\end{table*}

\begin{figure*}[t]
  \vskip 0.1in
  \begin{center}
    \includegraphics[width=\linewidth]{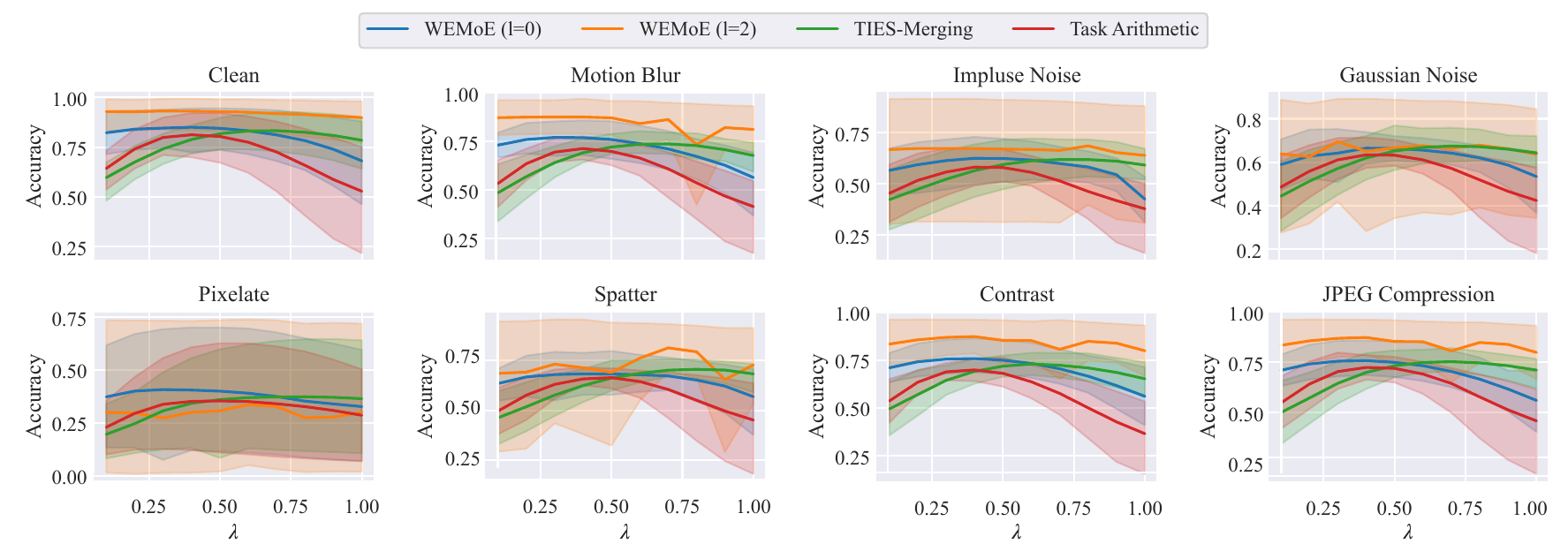}
  \end{center}
  \vskip -0.1in
  \caption{
    The results for robustness experiment on CLIP-ViT-B/32.
    The x-axis of each plot represents the scaling coefficient $\lambda$ of task vectors, while the y-axis shows the accuracy of the merged model on different merged tasks.
  }
  \label{fig:b32_robustness_lambda}
  \vskip 0.1in
\end{figure*}
\begin{figure}[htb]
  \vskip 0.2in
  \begin{center}
    \begin{subfigure}[b]{0.24\linewidth}
      \includegraphics[width=\linewidth,height=\linewidth]{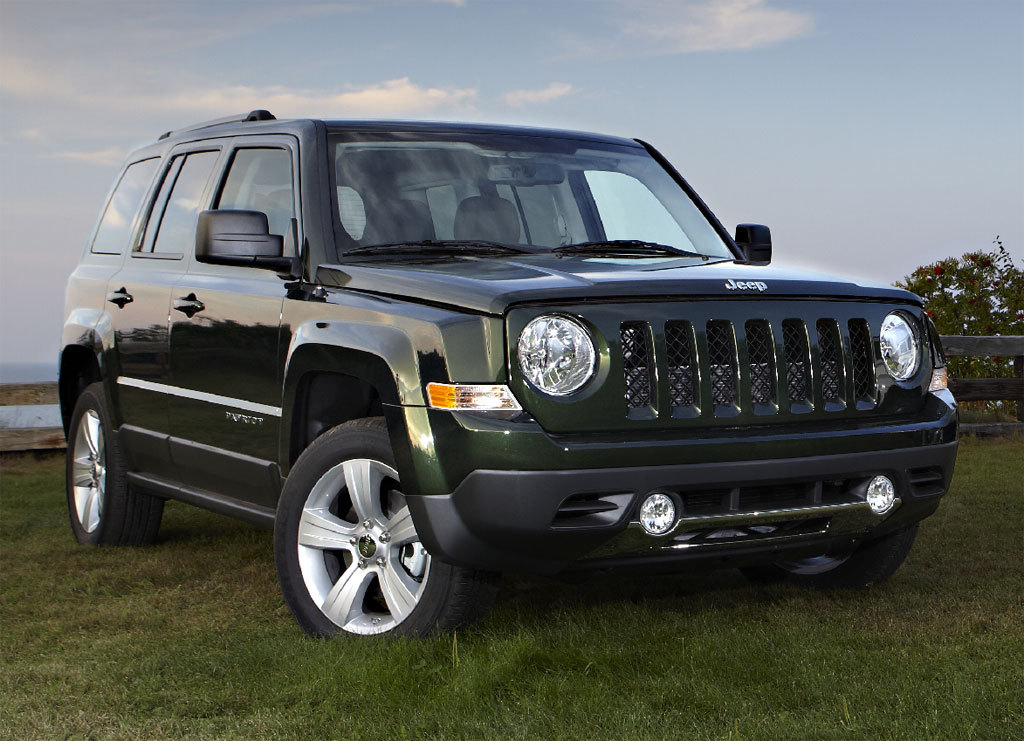}
      \caption{Clean}
    \end{subfigure}
    \begin{subfigure}[b]{0.24\linewidth}
      \includegraphics[width=\linewidth,height=\linewidth]{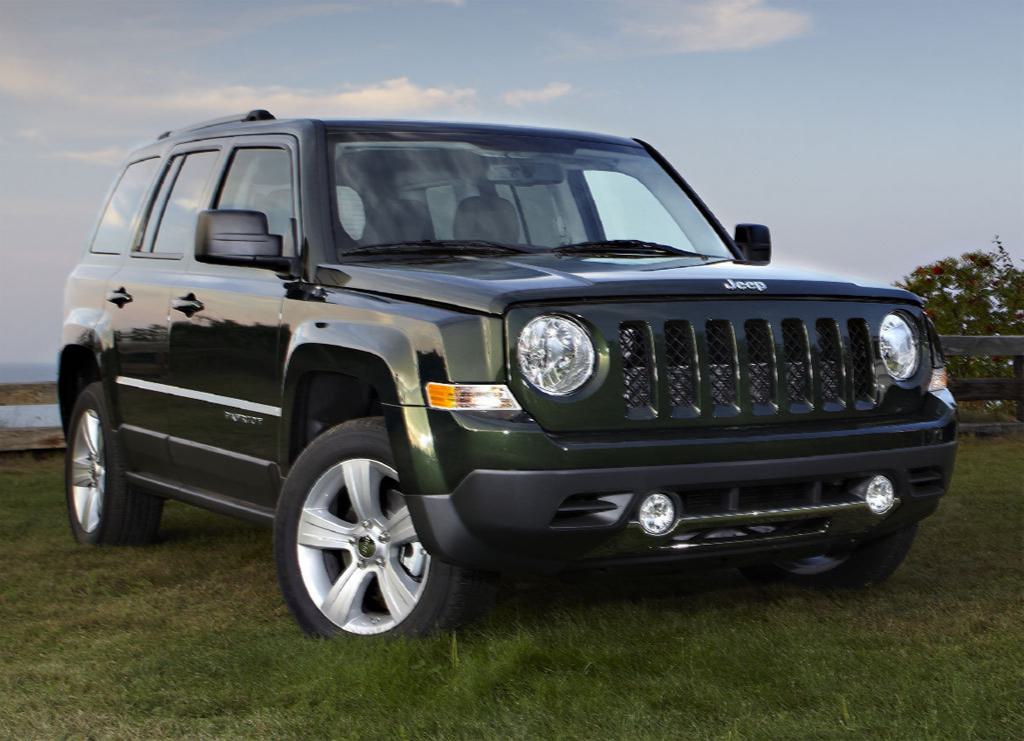}
      \caption{Motion}
    \end{subfigure}
    \begin{subfigure}[b]{0.24\linewidth}
      \includegraphics[width=\linewidth,height=\linewidth]{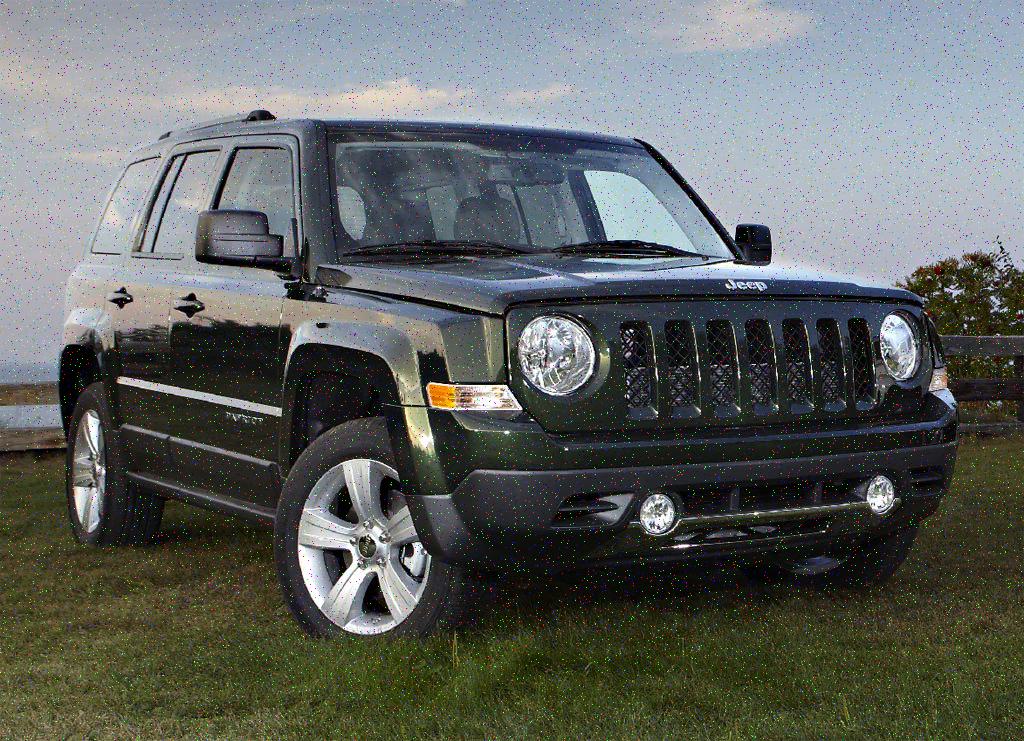}
      \caption{Impulse}
    \end{subfigure}
    \begin{subfigure}[b]{0.24\linewidth}
      \includegraphics[width=\linewidth,height=\linewidth]{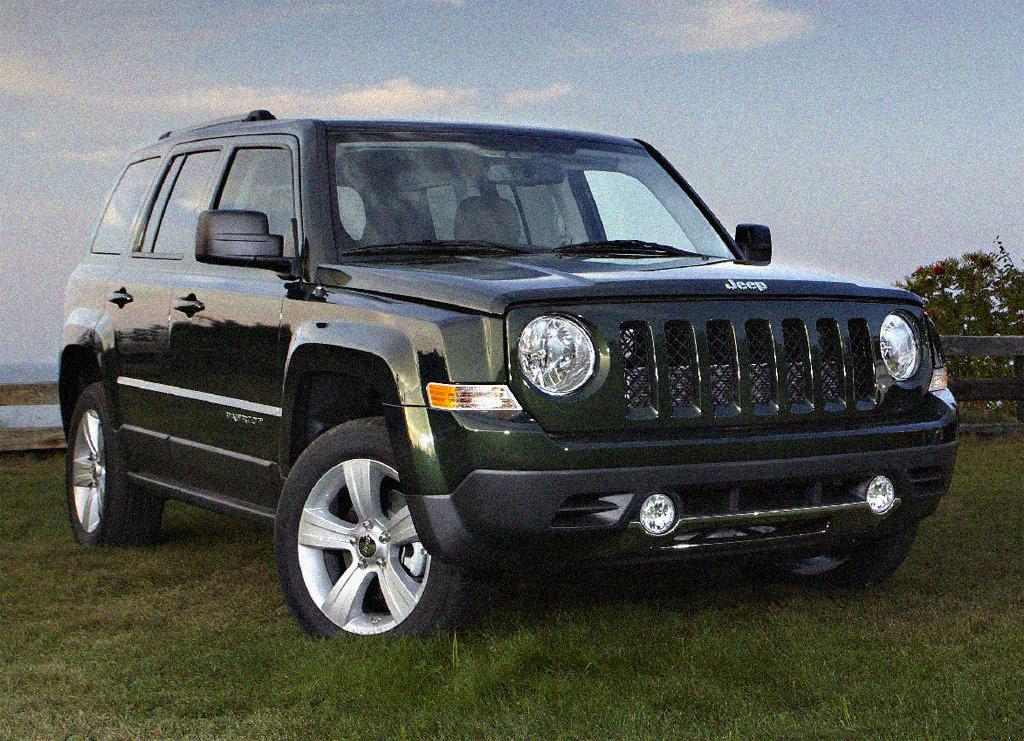}
      \caption{Gaussian}
    \end{subfigure}
    \begin{subfigure}[b]{0.24\linewidth}
      \includegraphics[width=\linewidth,height=\linewidth]{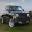}
      \caption{Pixelate}
    \end{subfigure}
    \begin{subfigure}[b]{0.24\linewidth}
      \includegraphics[width=\linewidth,height=\linewidth]{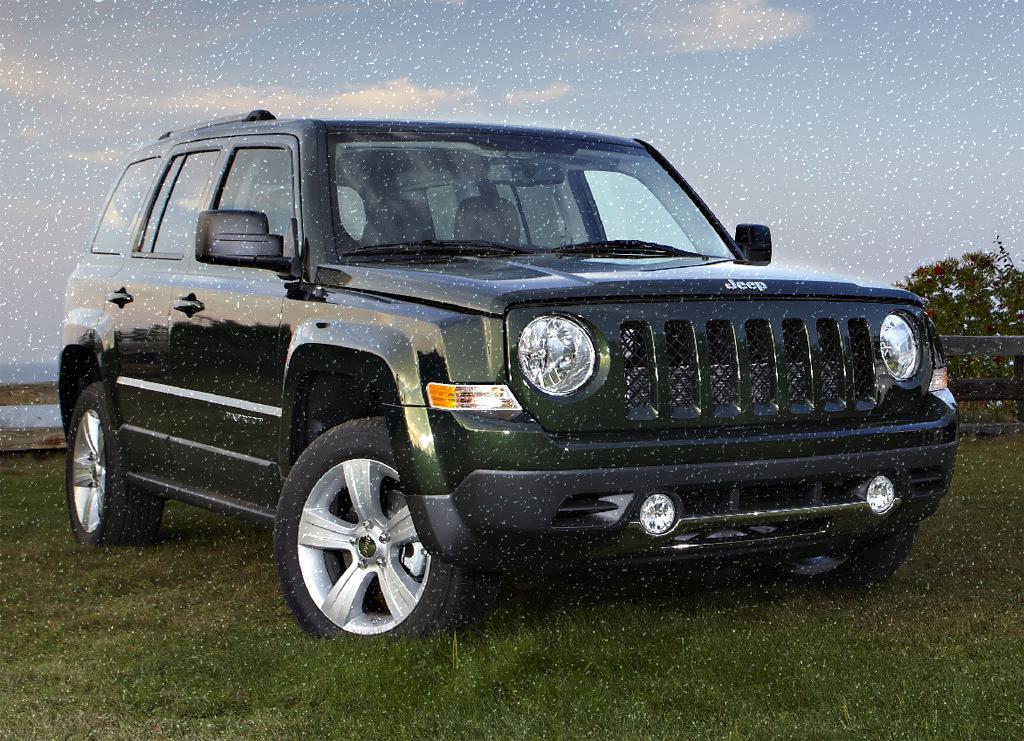}
      \caption{Spatter}
    \end{subfigure}
    \begin{subfigure}[b]{0.24\linewidth}
      \includegraphics[width=\linewidth,height=\linewidth]{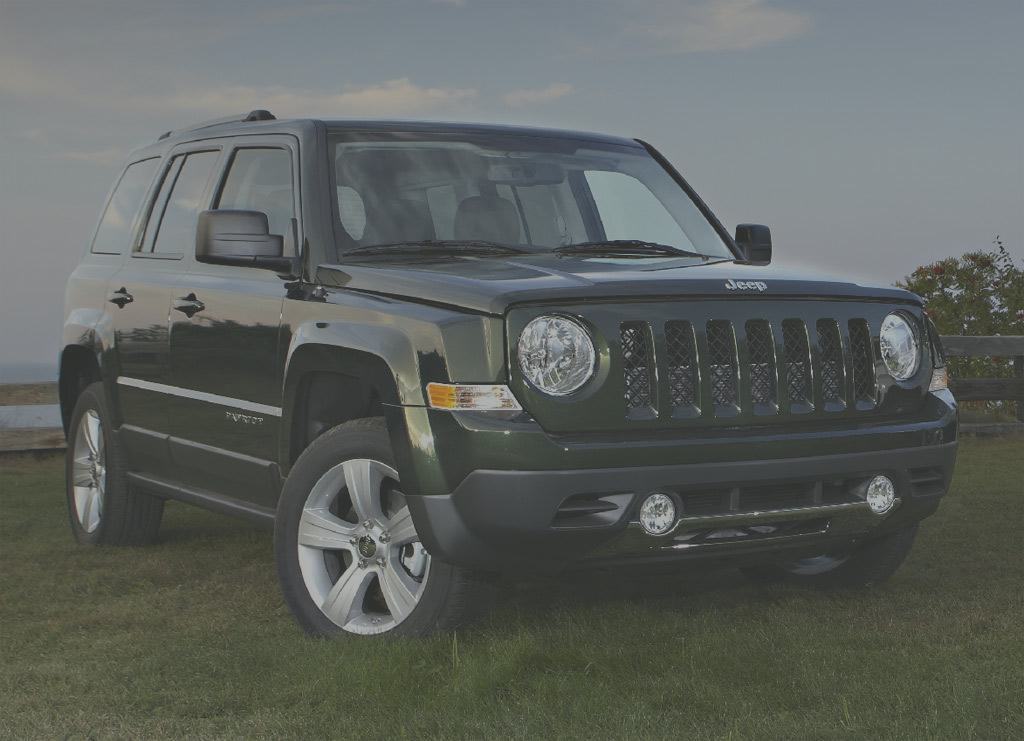}
      \caption{Contrast}
    \end{subfigure}
    \begin{subfigure}[b]{0.24\linewidth}
      \includegraphics[width=\linewidth,height=\linewidth]{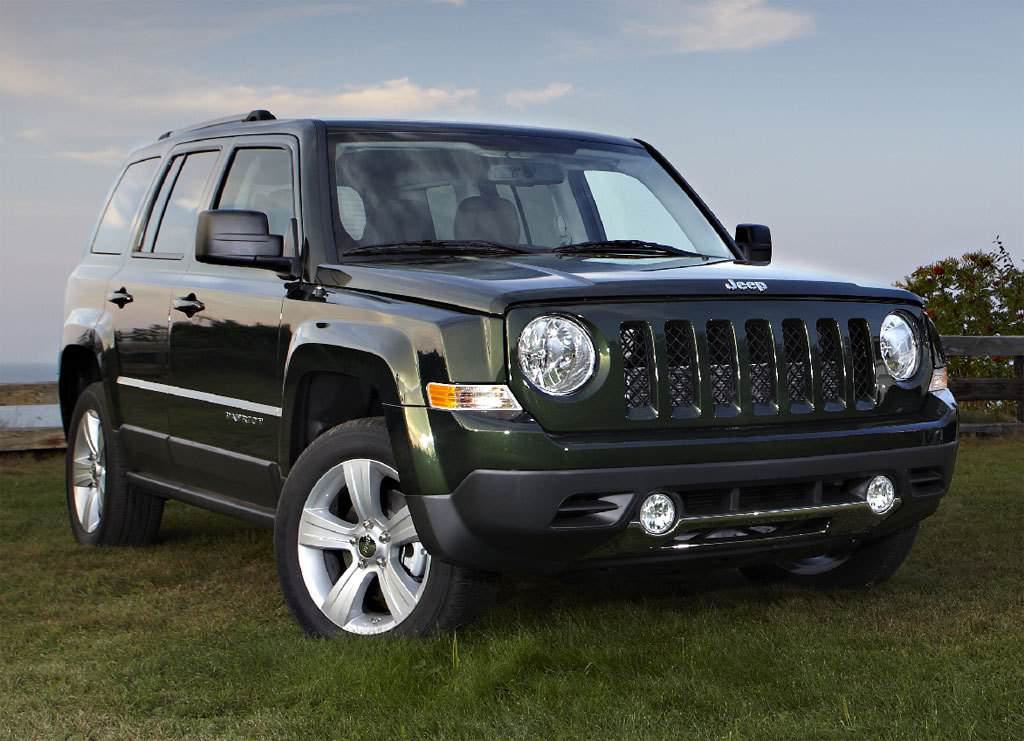}
      \caption{JPEG}
    \end{subfigure}
    \caption{
      Here are eight instances of distorted images, produced using the method suggested in~\citep{hendrycksBenchmarkingNeuralNetwork2019}.
    }
    \label{fig:distorted_images}
  \end{center}
  \vskip -0.2in
\end{figure}

\subsection{Experimental Setup}

We employ CLIP~\citep{radfordLearningTransferableVisual2021} as our pre-trained models, which are trained on a large-scale dataset with pairs of images and corresponding textual descriptions, and are able to perform open-vocabulary image classification.
Subsequently, we fine-tune the models on eight distinct image classification tasks, namely SUN397~\citep{xiao_sun_2010}, Stanford Cars~\citep{krause_3d_2013}, RESISC45~\citep{cheng_remote_2017}, EuroSAT~\citep{helber2018introducing}, SVHN~\citep{netzer_reading_2021}, GTSRB~\citep{stallkamp_man_2012}, MNIST~\citep{lecunGradientbasedLearningApplied1998}, and DTD~\citep{cimpoi_describing_2014}.
The evaluation of model performance is based on top-1 accuracy.

We compare our method with Fisher Merging~\citep{matenaMergingModelsFisherWeighted2022}, RegMean~\citep{jinDatalessKnowledgeFusion2023}, Task Arithmetic~\citep{ilharcoEditingModelsTask2023},  Ties-Merging~\citep{yadavResolvingInterferenceWhen2023}, and AdaMerging/Adamerging++~\citep{yangAdaMergingAdaptiveModel2023}.
For all methods, unless explicitly specified, we follow the configuration in \citep{yangAdaMergingAdaptiveModel2023} and initialize the scaling coefficient of the task vector, denoted as $\lambda$, to 0.3.
In Appendix~\ref{appendix:baseline_methods}, we provide a comparison of the baselines.

\subsection{Conventional Multi-Task Model Merging}

Firstly, we conduct conventional multi-task model merging experiments to evaluate the performance of our proposed method.
In Tables~\ref{table:parameter_counts} and \ref{table:parameter_counts_detailed}, we provides a detailed comparison of the parameter counts of the up-scaled models from eight tasks.
The models compared include CLIP-ViT-B/32, CLIP-ViT-B/16, and CLIP-ViT-L/14, with two different router depths ($l=0$ and $l=2$).
The experimental results are outlined in Table~\ref{table:multi-task_performance_clip-vit-b-32} and Table~\ref{table:multi-task_performance_clip-vit-l-14}, respectively.
We have the following key observations:

(1) The performance of fine-tuned models on downstream tasks is significantly better than that of the pre-trained model, indicating that the fine-tuned models have learned task-specific knowledge.
(2) For larger models, the fused performance tends to be better across all merging methods.
This is because the larger model scale introduces more redundancy among parameters.
Consequently, even simple averaging can achieve better performance on larger models.
(3) Our proposed approaches, WEMoE, consistently demonstrate the superiority of our approach over SOTA methods across the majority of tasks.
(4) In particular, WEMoE outperforms the traditional MTL baseline by 0.5\% and 0.1\% on average. This is surprising as the traditional MTL baseline is trained on all tasks simultaneously while merging methods are inaccessible to the training data.
This indicates that through our approach, we have successfully separated the task-specific knowledge from fine-tuned models.
On the other hand, the weight-ensembling MoE module successfully captures the relationship between the input features and knowledge combination, effectively alleviating the mutual influence of parameters, which significantly enhances the overall performance and flexibility of the model.

\textbf{Convergence and sensitivity analysis}.
Figure~\ref{fig:wemoe_fusion} shows the performance of the merged WEMoE models with varying number of steps.
In Figure~\ref{fig:b32_fusion}, we merge CLIP-ViT-B/32 models with different learning rate configurations.
We observe that the performance of the merged model shows an upward trend with an increase in the number of training steps, and it converges rapidly, reaching a high accuracy level in just 200 steps.
Furthermore, the influence of different learning rates is not significant, suggesting that our method is insensitive to the learning rate parameter. This is a desirable property as it reduces the need for hyperparameter tuning. In Figure~\ref{fig:clip_b32_and_l14_fusion}, we compare the performance of CLIP-ViT-B/32 and CLIP-ViT-L/14 models.

\begin{table*}[htb]
  \caption{Comparison of different up-scaling strategies on various tasks.}
  \label{table:upscaling_strategies}
  \centering
  \small
  \setlength{\tabcolsep}{1.5mm}
  \begin{tabular}{lccccccccc}
    \toprule
    \textbf{Method (1-layer router)} & \textbf{SUN397} & \textbf{Cars} & \textbf{RESISC45} & \textbf{EuroSAT} & \textbf{SVHN} & \textbf{GTSRB} & \textbf{MNIST} & \textbf{DTD}  & \textbf{Avg.} \\
    \midrule
    Entire Transformer Block         & 70.5            & 74.0          & 91.0              & 97.0             & 93.6          & 96.4           & 99.1           & 70.0          & 86.4          \\
    Attention + MLP (separately)     & \textbf{73.5}   & \textbf{78.4} & \textbf{94.0}     & 98.3             & \textbf{96.2} & 98.4           & \textbf{99.6}  & \textbf{75.6} & \textbf{89.2} \\
    MLP only                         & 73.2            & 76.7          & 93.8              & \textbf{98.6}    & 95.7          & \textbf{98.6}  & 99.5           & 74.5          & 88.3          \\
    \bottomrule
  \end{tabular}
\end{table*}

\textbf{Ablations of the up-scaling strategy}.
To further investigate the impact of the up-scaling strategy, we conduct experiments on CLIP-ViT-B/32.
We compare three different up-scaling strategies:
(1) Entire Transformer Block: The MoE up-scaling is applied on the entire Transformer block.
(2) Attention + MLP (separately): The MoE up-scaling is applied on the attention weights and MLP weights separately.
(3) MLP only: The MoE up-scaling is applied on the MLP weights only, as we have done in the main experiments.

In Table~\ref{table:upscaling_strategies}, we present the results of this ablation study.
When both attention and MLP weights are up-scaled separately, the performance is significantly improved across most tasks, surpassing the other two methods.
Recall that the decision to apply the routing network solely to the MLP weights was intentional, driven by the empirical finding that the MLP weights show less similarity between fine-tuned and pre-trained models than the attention weights (Section~\ref{subsection:revisiting_model_merge}).
This implies that MLPs may contain more task-specific information, making them an ideal target for the routing mechanism.
If the routing network were applied to both MLPs and attention weights, it would result in increased computational load and memory usage, and the marginal performance gain might not warrant the cost.

\subsection{Generalization and Robustness Evaluation}

When we engage in merging multi-task models, our primary objective is to enhance the model's performance on the seen task.
However, it is also crucial to explore the model's ability to generalize across diverse data distributions and assess the robustness of the merging algorithm when faced with shifts in test distribution. This refers to scenarios where the distribution of test data deviates from that of the training data, which is common in real-world applications.

\textbf{Generalization experiments.}
To assess the model's generalization, we selected two tasks from the eight downstream tasks as unseen tasks. We utilized the fine-tuned models on the remaining six tasks for merging, constructing a six-task model. This six-task model was then applied to the unseen tasks to evaluate the model's generalization capability.
We conducted experiments on both 0-layer routers and 2-layer routers, and the results are presented in Table~\ref{table:generalization_results_clip-vit-b-32}.

\textbf{Robustness experiments.}
To evaluate the robustness of the merging algorithm, we utilized the methods suggested in~\citep{hendrycksBenchmarkingNeuralNetwork2019} to generate distorted images from the clean test set.
We selected seven types of distortions, including motion blur, impulse noise, gaussian noise, pixelate, spatter, contrast, and JPEG compression.
We then merge and evaluate the performance of the merged models on the distorted images.
We compared routers of two different depths, see Eq.~(\ref{eq:leq0}) and Eq.~(\ref{eq:leq2}) for details.
We conduct experiments on CLIP-ViT-B/32 and CLIP-VIT-B/16, the results are presented in Table~\ref{table:abalation_data_distribution_vit_b_32} and Table~\ref{table:abalation_data_distribution_vit_b_16}, respectively.
Figure~\ref{fig:distorted_images} shows eight instances of distorted images.

Our experimental results show that across various methods, WEMoE ($l=2$) consistently achieves the highest performance on the clean test set and most of the distorted test sets.
This indicates that our method is effective in handling both clean and distorted data. The superior performance suggests its potential for robust multi-task merging.
However, in scenarios where there is a significant loss in image quality, such as pixelation, WEMoE ($l=2$) might overfit some specific tasks, resulting in a performance decline. In contrast, WEMoE ($l=0$) tends to exhibit more stability in performance under such conditions. This is attributed to its lower parameter count, making it less prone to overfitting.

In addition, we observe that the performance of WEMoE ($l=0$) is comparable to AdaMerging, recalling that when $l=0$, WEMoE is essentially equivalent to a partial implementation of AdaMerging, with the scaling factor $\lambda$ fixed at $0.3$ for the majority of the model, except for MLP modules.
This observation validates the hypothesis we presented in Section~\ref{subsection:revisiting_model_merge} regarding parameter similarity.

We also analyze the performance of our method with different scaling coefficients $\lambda$ of the task vector. The results are presented in Figure~\ref{fig:b32_robustness_lambda} and Figure~\ref{fig:b16_robustness_lambda}, respectively.
We observe that the performance of the merged model is relatively stable with respect to the scaling coefficient $\lambda$ of the task vector. This indicates that our method is also more robust to the scaling coefficient $\lambda$ of the task vector.

\subsection{Routing Analysis}

Here we perform a small analysis of the expert weighting by the router. We use the CLIP-ViT-B/32 model as an example, and the results are presented in Figure~\ref{fig:router_analysis}.
Our examination reveals a tendency of the router to allocate a greater weight to the task vector corresponding to the source task of the input sample.
This indicates that the routers are able to effectively identify the expert with the highest performance depending on the input feature, and assign a higher weight to it. Such behavior aligns with our expectations and intuition. Additional details can be found in Appendix~\ref{appendix:routing_analysis}.

\section{Related Work}
\label{section:related_work}

In this section, we review recent related works on multi-task model fusion and mixture of experts (MoE).

\textbf{Multi-Task Model Fusion}.
Weight interpolation proves to be a straightforward yet powerful strategy, facilitating scalable model fusion without the need for extensive computations~\citep{izmailovAveragingWeightsLeads2019,matenaMergingModelsFisherWeighted2022,wortsmanModelSoupsAveraging2022,kaddourStopWastingMy2022,ilharcoEditingModelsTask2023,yadavResolvingInterferenceWhen2023,yangAdaMergingAdaptiveModel2023,wuPiTuningTransferring2023}.
However, this strategy also poses challenges, particularly when merging models with diverse structures.

Mode connectivity has been uncovered by insights into the loss landscape~\citep{danielfreemanTopologyGeometryHalfrectified2017,nagarajanUniformConvergenceMay2019}.
This phenomenon reveals that different solutions can be connected by a pathway within the parameter space, maintaining low objective function values along the path, thus facilitating model fusion~\citep{draxlerEssentiallyNoBarriers2019,frankleLinearModeConnectivity2020,entezariRolePermutationInvariance2022}.
Different methods such as discovering simple linear paths~\citep{garipovLossSurfacesMode2018}, non-linear trajectories~\citep{tatroOptimizingModeConnectivity2020}, or mappings within a lower dimension~\citep{yunisConvexityLinearMode2022,bentonLossSurfaceSimplexes2021} have been proposed to leverage this concept.

Alignment emerges as another effective series of works. This approach involves aligning and interpolating corresponding components from various models to mitigate disparities~\citep{liConvergentLearningDifferent2016,tatroOptimizingModeConnectivity2020}. Techniques include matching activations or weights~\citep{georgestoicaZipItMergingModels2023,jinDatalessKnowledgeFusion2023}, employing graph matching for channel-wise alignment~\citep{liuDeepNeuralNetwork2022a}, or exploiting the principle of permutation invariance~\citep{ainsworthGitReBasinMerging2023}.

\textbf{Mixture of Experts}.
The Mixture of Experts (MoE) model, first introduced by~\cite{jacobsAdaptiveMixturesLocal1991}, is a machine learning technique that involves training multiple models, each of which specializes in a different part of the input space. Over the years, MoE has garnered considerable attention~\citep{jiangMixtralExperts2024,daiDeepSeekMoEUltimateExpert}.
Much innovation revolves around the design of more efficient routers.
For instance, the Switch Transformer~\citep{fedusSwitchTransformersScaling2022} simplifies the selection process by choosing only the top expert per token, showcasing superior scalability compared to prior approaches. Base Layers~\citep{lewisBASELayersSimplifying2021} introduces a linear assignment that optimizes token-expert affinities, ensuring an equal distribution of tokens among experts.
In addition to methods that involve routers selecting experts, there are alternative approaches such as allowing each expert to choose tokens~\citep{zhouMixtureofExpertsExpertChoice2022}.
For a comprehensive review of MoE, readers can refer to~\citep{fedusReviewSparseExpert2022}.

\section{Conclusion}
\label{section:conclusion}

In this paper, we propose a novel method to merge Transformer-based vision models from different tasks. 
Our method is effective in transferring knowledge from various task-specific fine-tuned models and constructing a unified multi-task model. We propose a novel Weight-Ensembling MoE (WEMoE) module, which can dynamically integrate shared and task-specific knowledge based on the input sample. We conduct extensive experiments, the experimental results demonstrate the effectiveness of our method and provide a comprehensive understanding of our method. 

In the future, we plan to explore the potential of our method in other scenarios, such as merging transformers from different modality. We also plan to investigate the possibility of applying our method to other architectures, such as CNNs. 
It's also interesting to combine our method with parameter-efficient fine-tuning methods such as Adapter tuning~\citep{houlsbyParameterEfficientTransferLearning2019} and LoRA~\citep{huLoRALowRankAdaptation2021} to further improve the efficiency of our method.
\FloatBarrier

\section*{Acknowledgements}

This work is supported in part by STI 2030—Major Projects (No. 2021ZD0201405), the National Natural Science Foundation of China (Grant No. 62276195 and U23A20318), the Special Fund of Hubei Luojia Laboratory under Grant 220100014, and the Fundamental Research Funds for the Central Universities (No. 2042024kf0039).
Dr Tao's research is partially supported by NTU RSR and Start Up Grants.

\section*{Impact Statement}

This work presents a novel method for merging Transformer-based models, aiming to advance the field of machine learning, particularly in the area of multi-task learning. The proposed method has the potential to significantly improve the efficiency and scalability of multi-task learning systems, thereby enabling more effective knowledge transfer and utilization.
The societal implications of this work are manifold. On the positive side, the proposed method could lead to more efficient and effective machine learning systems, which could in turn lead to advancements in various fields where machine learning is applied, such as healthcare, education, and technology. This could potentially result in improved services and products, benefiting society at large.

\bibliography{references}
\bibliographystyle{icml2024}

\clearpage
\appendix
\onecolumn

The appendix is organized into several sections, each providing additional insights and details related to different aspects of the main work.

\startcontents[sections]  %
\printcontents[sections]{}{1}{\setcounter{tocdepth}{2}}  %
\vskip 0.2in
\hrule

\section{Loss Landscape Visualization}
\label{appendix:loss}

\begin{figure}[ht]
  \vskip 0.1in
  \begin{center}
    \begin{subfigure}{0.3\linewidth}
      \centering
      \includegraphics[width=0.88\linewidth]{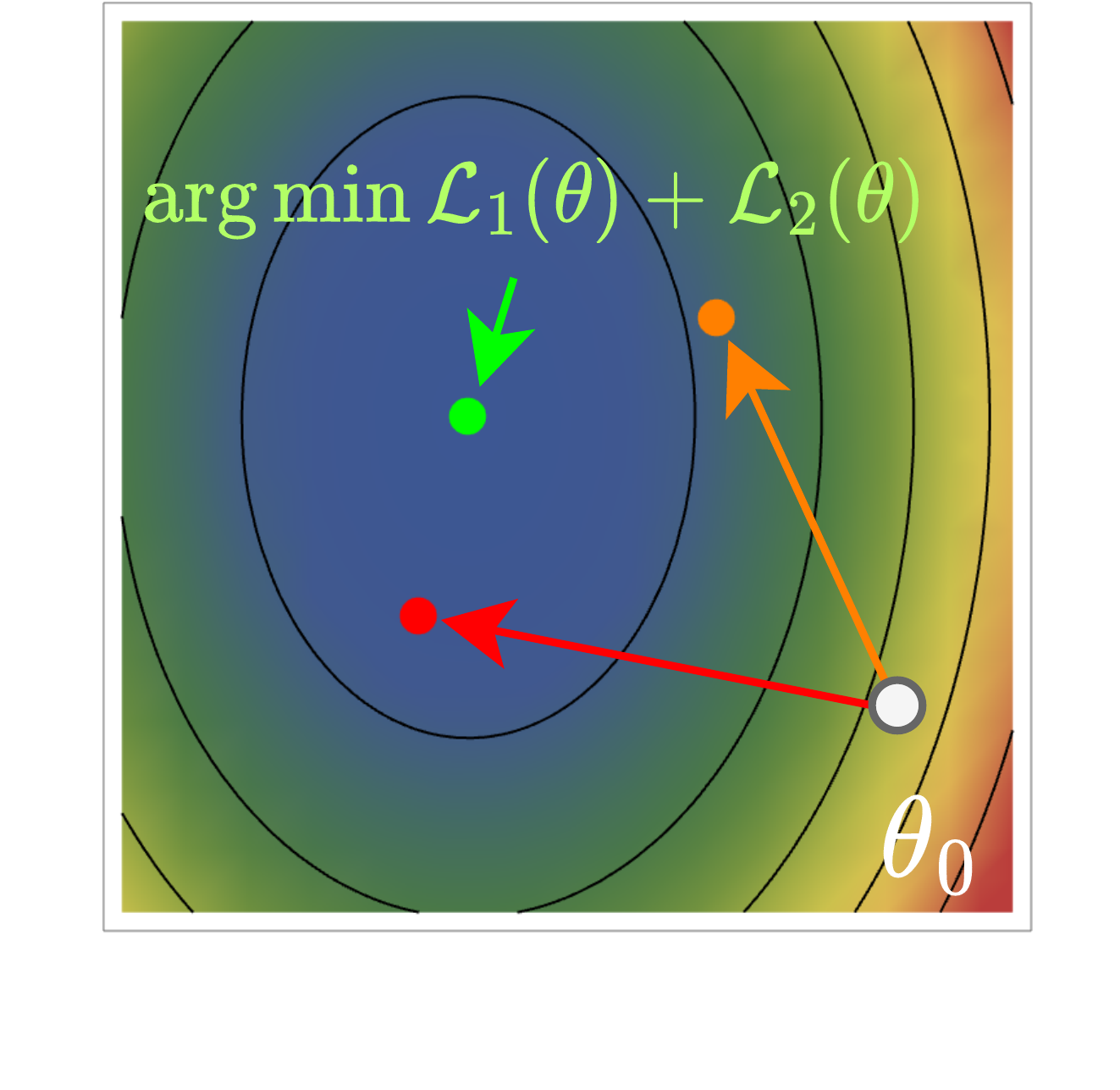}
      \caption{$\mathcal{L}_1 + \mathcal{L}_2$}
    \end{subfigure}
    \begin{subfigure}{0.3\linewidth}
      \centering
      \includegraphics[width=\linewidth]{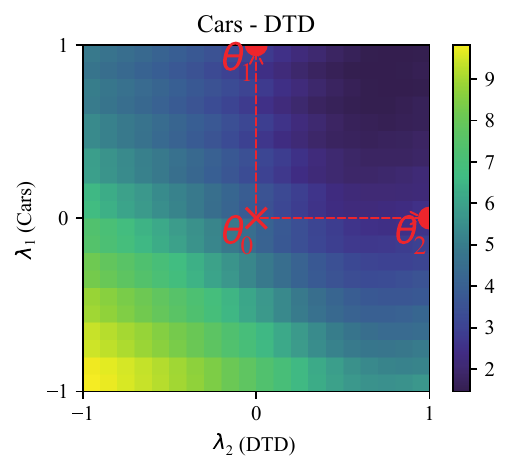}
      \caption{Cars-DTD}
    \end{subfigure}
    \begin{subfigure}{0.3\linewidth}
      \centering
      \includegraphics[width=\linewidth]{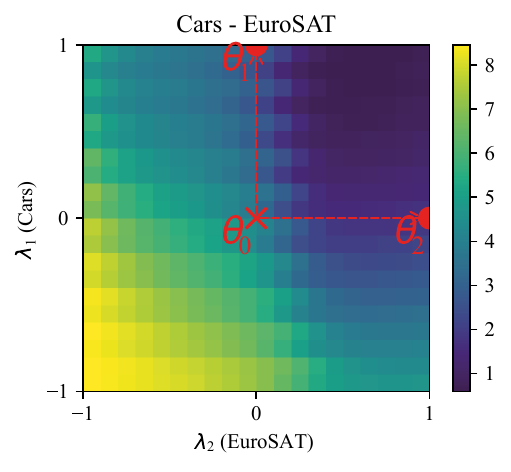}
      \caption{Cars-EuroSAT}
    \end{subfigure}
    \begin{subfigure}{0.3\linewidth}
      \centering
      \includegraphics[width=\linewidth]{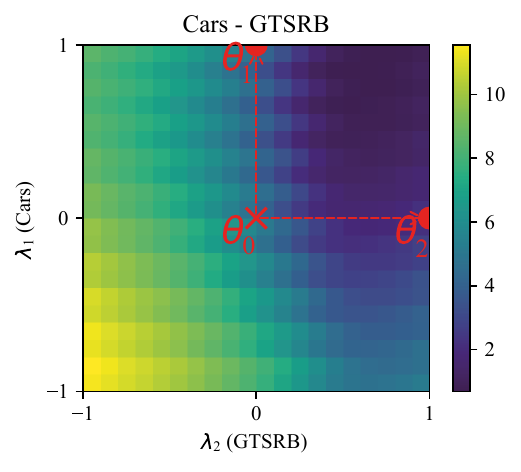}
      \caption{Cars-GTSRB}
    \end{subfigure}
    \begin{subfigure}{0.3\linewidth}
      \centering
      \includegraphics[width=\linewidth]{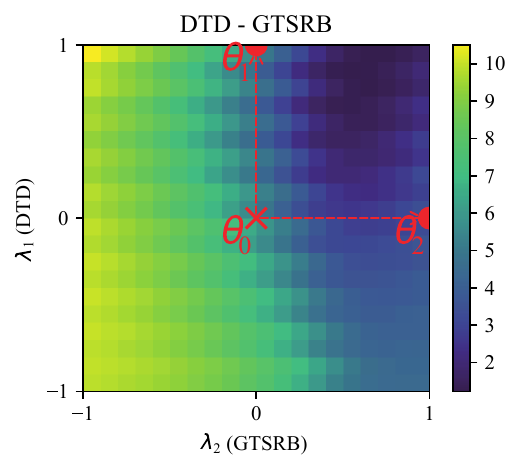}
      \caption{DTD-GTSRB}
    \end{subfigure}
    \begin{subfigure}{0.3\linewidth}
      \centering
      \includegraphics[width=\linewidth]{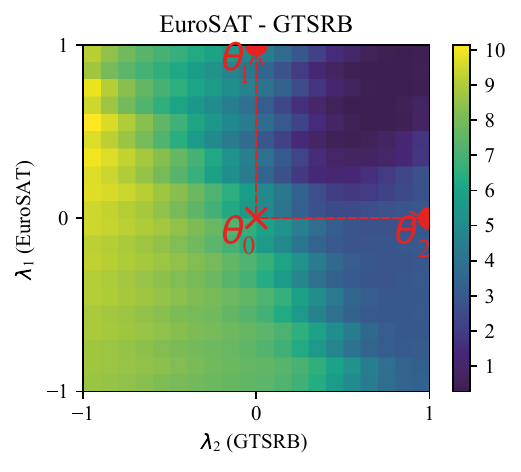}
      \caption{EuroSAT-GTSRB}
    \end{subfigure}
  \end{center}
  \caption{
    Visualization of the joint loss $\mathcal{L}_1 + \mathcal{L}_2$ and five task pairs for CLIP-ViT-B/32 in the loss landscape.
    We perform interpolations between pre-trained weights and two fine-tuned weights in the weight space on a 2D plane using the formula $\theta=\theta_0 + \lambda_1 \tau_1 + \lambda_2 \tau_2$, where $\theta_0$ represents pre-trained weights, $\tau_i=\theta_i -\theta_0$  are two task vectors with $\lambda_i$ in the range [-1, 1]. 
  }
  \label{fig:loss_landscapes_examples}
\end{figure}

In Section~\ref{subsection:revisiting_model_merge}, we highlight the inherent limitation of a static solution for multi-task model merging, emphasizing its potential to result in suboptimal performance on individual tasks. To further illustrate this point, we present the visualization of the loss landscape for the joint loss $\mathcal{L}_1 + \mathcal{L}_2$ and five task pairs using CLIP-ViT-B/32 in Figure~\ref{fig:loss_landscapes_examples}. The visualization involves interpolations between pre-trained weights and two fine-tuned weights in the weight space on a 2D plane, expressed as $\theta = \theta_0 + \lambda_1 \tau_1 + \lambda_2 \tau_2$, where $\theta_0$ represents pre-trained weights, and $\tau_i = \theta_i - \theta_0$ are two task vectors with $\lambda_i$ in the range [-1, 1]. The resulting heatmaps demonstrate that task-specific models fine-tuned from the same pre-trained model share the same loss basin when evaluated on the joint task, yet none of them attains the global optimum.

\section{Multi-Task Model Fusion}
\label{appendix:multi_task_model_fusion}

\subsection{Baseline methods}
\label{appendix:baseline_methods}

\begin{table}[h]
  \caption{Comparison of different methods and their data access requirements.}
  \label{table:method_comparison}
  \centering
  \small
  \setlength{\tabcolsep}{5pt}
  \begin{tabular}{lccc}
    \midrule
    \textbf{Method}                                              & \textbf{Labeled tasks data} & \textbf{Validation data (labeled)} & \textbf{Test time adaptation} \\
    \midrule
    Fisher Merging~\citep{matenaMergingModelsFisherWeighted2022} & Yes                         & No                                 & No                            \\
    RegMean~\citep{jinDatalessKnowledgeFusion2023}               & Yes                         & No                                 & No                            \\
    Task Arithmetic~\citep{ilharcoEditingModelsTask2023}         & No                          & Yes                                & No                            \\
    Ties-Merging~\citep{yadavResolvingInterferenceWhen2023}      & No                          & Yes                                & No                            \\
    AdaMerging~\citep{yangAdaMergingAdaptiveModel2023}           & No                          & No                                 & Yes                           \\
    \textbf{Ours}                                                & No                          & No                                 & Yes                           \\
    \bottomrule
  \end{tabular}
\end{table}

Here we provide a summary of the baseline methods used in the experiments conducted in this study.

\begin{itemize}
  \item \textbf{Simple Weight Average}:  This method involves taking the average of the weights of models that have been fine-tuned on different tasks. It is sometimes referred to as ModelSoups in relevant literature. When applied to fully fine-tuned models, the weights of the models are directly averaged. This is mathematically represented as $\theta = {1/n \sum_{i=1}^{n} \theta_i}$, where $\theta$ represents the average weight, $n$ is the total number of models, and $\theta_i$ is the weight of the $i$-th model.
  \item \textbf{Task Arithmetic}:  This method involves calculating a task vector for each task and then adding these vectors together to create a multi-task vector. This multi-task vector is then scaled by a coefficient $\lambda$ and added to the initial parameters of the pre-trained model to create a multi-task model. The formula for this is $\theta = \theta_0 + \lambda \sum_{i} (\theta_i - \theta_0)$, where $\theta_0$ is the initial parameter of the pre-trained model, $\theta_i$ is the task vector for the $i$-th task, and $\lambda$ is a hyperparameter chosen based on the model's performance on a validation set. In this study, $\lambda$ was set to $0.3$.
  \item \textbf{Ties-Merging}: This algorithm follows three steps (trim, elect sign of parameters, and disjoint merge) to obtain a merged task vector $\nu$. Given the final merged task vector $\tau$, the final model is chosen in a similar way as task arithmetic, i.e. $\theta = \theta_0 + \lambda \tau$, where $\lambda$ is a hyperparameter that the best-performing model is chosen on the validation set. In our study, $\lambda$ is chosen to be $0.3$.
  \item \textbf{Fisher Merging}: This method requires access to some labeled data from all tasks in order to estimate the Fisher information matrix. The Fisher information matrix is then used to assess the importance of each task and to merge the models by computing a weighted average of the models' weights.
  \item \textbf{RegMean}: This method also requires access to some labeled data from all tasks, but it is used to compute the Gram matrix. The Gram matrix is a matrix that represents the inner products of vectors in a set.
  \item \textbf{AdaMerging}: AdaMerging is an adaptive model merging method where it autonomously learns the coefficients for merging either on a task-wise or layer-wise basis, using entropy minimization on unlabeled test samples as a surrogate objective function to refine the merging coefficients.
        \begin{itemize}
          \item The task-wise AdaMerging is formulated as $\theta = \theta_0 + \sum_{i=1}^{n} \lambda_i \tau_i$ where $\lambda_k$ is the merging coefficient for the $k$-th task and $\tau_k$ is the task vector for the $k$-th task.
          \item The layer-wise AdaMerging is formulated as $\theta^l = \theta_0^l + \sum_{i=1}^{n} \lambda^{l}_{i} \tau^{l}_{i}$. Where the superscript $l$ denotes the layer index.
        \end{itemize}
\end{itemize}

\subsection{Image Classification Tasks}

\begin{table}[h]
  \caption{Comparison of parameter counts and reductions in WEMoE models with varying tasks.}
  \label{table:parameter_counts_detailed}
  \centering
  \small
  \begin{tabular}{lccc}
    \toprule
    \textbf{Method}          & \textbf{Trainable Parameters} & \textbf{Total Parameters} & \textbf{Parameters Reduced by Merging} \\
    \midrule
    Single Pre-trained       & 113.45M (100\%)               & 113.45M                   & -                                      \\
    WEMoE (2-layer, 2 tasks) & 7.11M (3.04\%)                & 233.89M                   & -6.99M                                 \\
    WEMoE (2-layer, 3 tasks) & 7.11M (2.45\%)                & 290.57M                   & 49.78M                                 \\
    WEMoE (2-layer, 4 tasks) & 7.12M (2.02\%)                & 347.25M                   & 106.55M                                \\
    WEMoE (2-layer, 5 tasks) & 7.13M (1.77\%)                & 403.93M                   & 163.32M                                \\
    WEMoE (2-layer, 6 tasks) & 7.14M (1.55\%)                & 460.61M                   & 220.09M                                \\
    WEMoE (2-layer, 7 tasks) & 7.15M (1.38\%)                & 517.28M                   & 276.87M                                \\
    WEMoE (2-layer, 8 tasks) & 7.16M (1.25\%)                & 573.96M                   & 333.64M                                \\
    \bottomrule
  \end{tabular}
\end{table}

\begin{figure}[ht]
  \begin{center}
    \begin{subfigure}{0.49\linewidth}
      \centering
      \includegraphics[width=\linewidth]{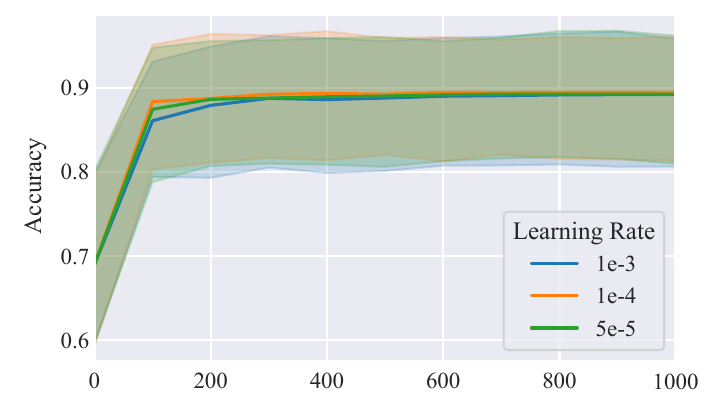}
      \caption{Learning rate comparison.}
      \label{fig:b32_fusion_large}
    \end{subfigure}
    \begin{subfigure}{0.49\linewidth}
      \centering
      \includegraphics[width=\linewidth]{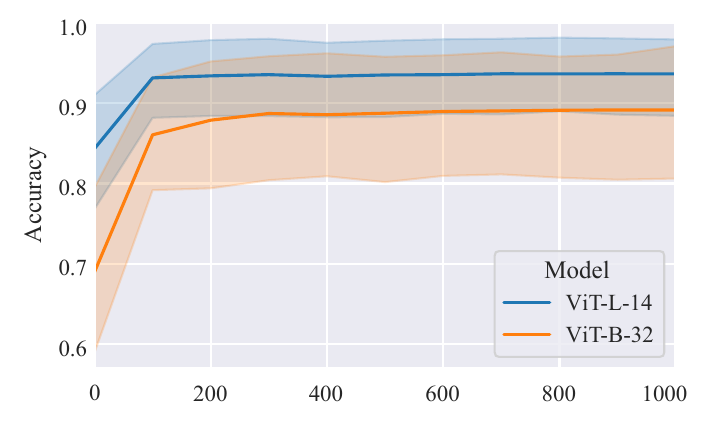}
      \caption{Model comparison.}
      \label{fig:clip_b32_and_l14_fusion_large}
    \end{subfigure}
    \caption{The performance of the merged models with a varying number of steps.
      (a) CLIP-ViT-B/32 model with different learning rates.
      (b) Comparison of CLIP-ViT-B/32 and CLIP-ViT-L/14.
    }
    \label{fig:wemoe_fusion_large}
  \end{center}
  \vskip -0.1in
\end{figure}

In this section, we conduct experiments to validate the effectiveness of our method.
We merge the CLIP-ViT-B/32 and CLIP-ViT-L/14 models on all eight tasks.
In Table~\ref{table:parameter_counts_detailed}, we compare the parameter counts and reductions in WEMoE models with varying tasks.
Where the proposed method has an MoE structure on MLP weights, which introduces additional parameters compared to a single pre-trained model (the MLPs have about 60\% of the parameters in a Transformer-based model). This is an inherent limitation of all mixture-of-experts based methods. It's an interesting direction to explore techniques to reduce memory overhead in future work, such as using a low-rank approximation or sparse structure.

Tables~\ref{table:multi-task_performance_clip-vit-b-32} and~\ref{table:multi-task_performance_clip-vit-l-14} present a comprehensive comparison of multi-task performance for CLIP-ViT-B/32 and CLIP-ViT-L/14 models across eight distinct tasks. The table includes results for various merging methods, such as traditional multi-task learning (MTL), Weight Averaging, Fisher Merging~\citep{matenaMergingModelsFisherWeighted2022}, RegMean~\citep{jinDatalessKnowledgeFusion2023}, Task Arithmetic~\citep{ilharcoEditingModelsTask2023}, Ties-Merging~\citep{yadavResolvingInterferenceWhen2023}, AdaMerging/AdaMerging++ (task-wise and layer-wise)~\citep{yangAdaMergingAdaptiveModel2023}, and our proposed WEMoE method.
Additionally, individual task performance, traditional MTL, and several model fusion methods are evaluated to provide a thorough understanding of the effectiveness of each approach.

We have the following key observations, which showcase WEMoE's effectiveness in multi-task model fusion:
\begin{enumerate}
  \item The performance of fine-tuned models on downstream tasks is significantly better than that of the pre-trained model, indicating that the fine-tuned models have learned task-specific knowledge.
  \item For larger models, the fused performance tends to be better across all merging methods.
        This is because the larger model scale introduces more redundancy among parameters.
        Consequently, even simple averaging can achieve better performance on larger models.
  \item Our proposed approach, WEMoE, consistently demonstrates the superiority of our approach over SOTA methods across the majority of tasks.
  \item In particular, WEMoE outperforms the traditional MTL baseline by 0.5\% and 0.1\% on average. This is surprising as the traditional MTL baseline is trained on all tasks simultaneously while merging methods are inaccessible to the training data.
\end{enumerate}
These observations indicate that through our approach, we have successfully separated the task-specific knowledge from fine-tuned models.
On the other hand, the weight-ensembling MoE module successfully captures the relationship between the input features and knowledge combination, effectively alleviating the mutual influence of parameters, which significantly enhances the overall performance and flexibility of the model.

Figure~\ref{fig:wemoe_fusion_large} shows the performance of the merged WEMoE models with varying number of steps.
In Figure~\ref{fig:b32_fusion_large}, we merge CLIP-ViT-B/32 models with different learning rate configurations.
We observe that the performance of the merged model shows an upward trend with an increase in the number of training steps, and it converges rapidly, reaching a high accuracy level in just 200 steps.
Furthermore, the influence of different learning rates is not significant, suggesting that our method is insensitive to the learning rate parameter. This is a desirable property as it reduces the need for hyperparameter tuning. In Figure~\ref{fig:clip_b32_and_l14_fusion_large}, we compare the performance of CLIP-ViT-B/32 and CLIP-ViT-L/14 models.

\subsection{Ablations of Router Ddepth}
\label{appendix:ablations_of_router_depth}

\begin{table}[h]
  \caption{
    Parameter comparison of WEMoE (1-layer) and WEMoE (2-layer) on CLIP-ViT-B/32 models.
    We add AdaMerging as a baseline for comparison.
  }
  \label{table:router_depth_parameter_count}
  \centering
  \small
  \begin{tabular}{lc}
    \toprule
    \textbf{Method}         & \textbf{Number of Trainable Parameters} \\
    \midrule
    AdaMerging (layer-wise) & 1.3K                                    \\
    WEMoE (1-layer)         & 73.8K(0.01\%)                           \\
    WEMoE (2-layer)         & 7.16M(1.25\%)                           \\
    \bottomrule
  \end{tabular}
\end{table}

\begin{table}[h]
  \caption{
    Ablation study of the router depth on the performance of the up-scaled CLIP-ViT-B/32 models.
    We add AdaMerging as a baseline for comparison.
  }
  \label{table:router_depth}
  \begin{center}
    \setlength{\tabcolsep}{3pt}
    \begin{small}
      \begin{sc}
        \begin{tabular}{lccccccccc}
          \toprule
          \textbf{Method}         & \textbf{SUN397} & \textbf{Cars} & \textbf{RESISC45} & \textbf{EuroSAT} & \textbf{SVHN} & \textbf{GRSRB} & \textbf{MNIST} & \textbf{DTD} & \textbf{Avg.} \\
          \midrule
          AdaMerging (layer-wise) & 66.6            & 68.3          & 82.4              & 92.5             & 86.5          & 93.7           & 97.7           & 61.1         & 80.9          \\
          WEMoE (1-layer)         & 73.2            & 76.7          & 93.8              & 98.6             & 95.7          & 98.6           & 99.5           & 74.5         & 88.3          \\
          WEMoE (2-layer)         & 74.1            & 77.4          & 93.7              & 99.1             & 96.2          & 98.9           & 99.6           & 76.4         & 89.4          \\
          \bottomrule
        \end{tabular}
      \end{sc}
    \end{small}
  \end{center}
\end{table}

To explore the influence of router depth on the performance of the scaled-up model, we perform an ablation study where the router depth is varied.
Before we delve into the experimental results, let's clarify the router depth in our WEMoE model, as introduced in Section~\ref{subsection:weight_ensembling_moe_module}.
In our WEMoE modules, the router is implemented as a multi-layer perceptron (MLP).
\begin{itemize}
  \item WEMoE (0-layer) functions as a bias-only model, representing a special case of an MLP with no hidden layers. It generates a constant routing weight for all inputs, captured by the formula as follows
        \begin{equation}
          r(h) = b_0,
        \end{equation}
        indicating that it does not adjust based on the input.
        When we only up-scale the MLP modules of the vision Transformers to MoE modules, WEMoE (0-layer) can be considered as a partial implementation of AdaMerging. Add when we up-scale the vision Transformers layer-wisely, WEMoE (0-layer) can be considered equivalent to AdaMerging.
        For WEMoE (0-layer), the MoE modules can be unloaded, thus no additional parameters and inference cost are introduced.
  \item For WEMoE (1-layer), each router is a one-layer MLP that takes the input sample $h$ and outputs the routing weight $r(h)$, which is adaptive to the input. The routing weight is calculated as follows
        \begin{equation}
          r(h) = W_1 h + b_1.
        \end{equation}
  \item For WEMoE (2-layer), each router is a two-layer MLP and the routing weight is calculated as follows
        \begin{equation}
          r(h) = W_2 ReLU(W_1 h + b_1) + b_2.
        \end{equation}
\end{itemize}

In Tables~\ref{table:generalization_results_clip-vit-b-32} and \ref{table:abalation_data_distribution_vit_b_32}, we note a substantial performance improvement when comparing WEMoE (2-layer) to WEMoE (0-layer).
While it may appear that the auxiliary parameters are the primary contributors to the final performance, rather than the proposed MoE design, this is not the case.
In fact, the key innovation of our approach lies in the proposed MoE design, which offers the ability to dynamically integrate shared and task-specific knowledge based on input, i.e. the data adaptability of the routing mechanism.

In Tables~\ref{table:router_depth_parameter_count} and \ref{table:router_depth}, we present additional findings to support our argument. We compare the number of trainable parameters and performance between WEMoE (1-layer) and WEMoE (2-layer). The data reveal that WEMoE (1-layer) possesses 73.8K trainable parameters, which constitute only 0.01\% of the total parameters in the merged model. Notably, the performance of WEMoE (1-layer) is significantly better than AdaMerging and nearly matches that of WEMoE (2-layer) across all tasks. This evidence underscores our claim that the MoE design is crucial for performance enhancement.

\section{Robustness to Out-of-Distribution Data}

\begin{figure*}[h]
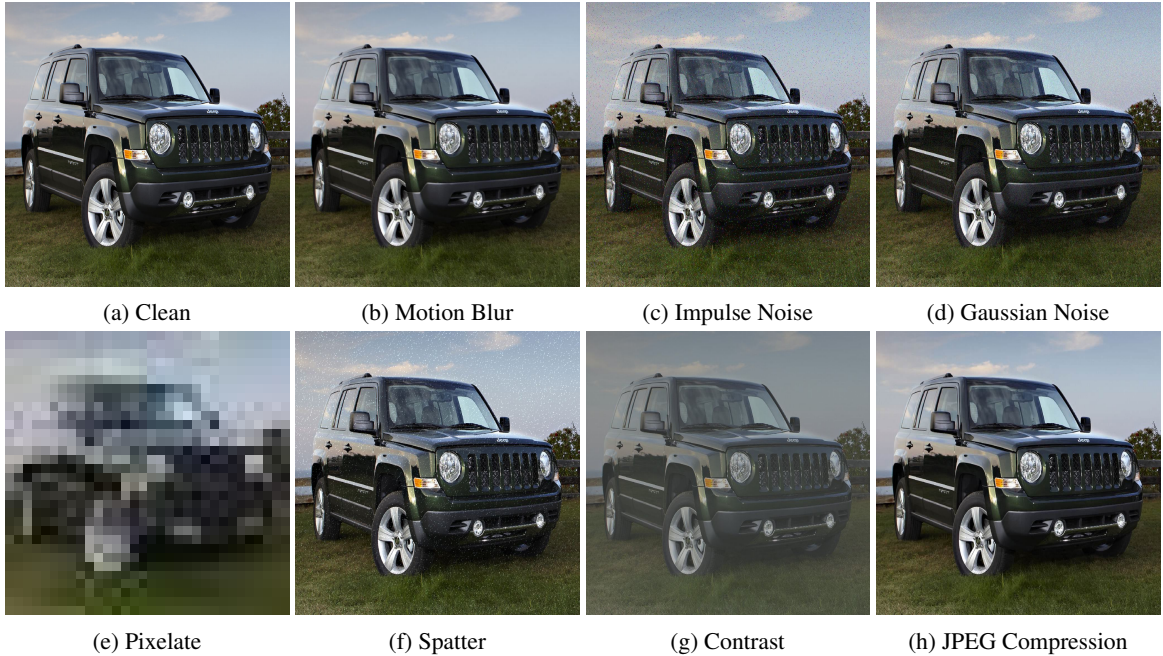

  \vskip 0.2in
  \begin{center}
    \begin{subfigure}[b]{0.22\linewidth}
      \includegraphics[width=\linewidth,height=\linewidth]{figure/clean-00003.jpg}
      \caption{Clean}
    \end{subfigure}
    \begin{subfigure}[b]{0.22\linewidth}
      \includegraphics[width=\linewidth,height=\linewidth]{figure/motion_blur-00003.jpg}
      \caption{Motion Blur}
    \end{subfigure}
    \begin{subfigure}[b]{0.22\linewidth}
      \includegraphics[width=\linewidth,height=\linewidth]{figure/impulse_noise-00003.jpg}
      \caption{Impulse Noise}
    \end{subfigure}
    \begin{subfigure}[b]{0.22\linewidth}
      \includegraphics[width=\linewidth,height=\linewidth]{figure/gaussian_noise-00003.jpg}
      \caption{Gaussian Noise}
    \end{subfigure}
    \begin{subfigure}[b]{0.22\linewidth}
      \includegraphics[width=\linewidth,height=\linewidth]{figure/pixelate-00003.jpg}
      \caption{Pixelate}
    \end{subfigure}
    \begin{subfigure}[b]{0.22\linewidth}
      \includegraphics[width=\linewidth,height=\linewidth]{figure/spatter-00003.jpg}
      \caption{Spatter}
    \end{subfigure}
    \begin{subfigure}[b]{0.22\linewidth}
      \includegraphics[width=\linewidth,height=\linewidth]{figure/contrast-00003.jpg}
      \caption{Contrast}
    \end{subfigure}
    \begin{subfigure}[b]{0.22\linewidth}
      \includegraphics[width=\linewidth,height=\linewidth]{figure/jpeg_compression-00003.jpg}
      \caption{JPEG Compression}
    \end{subfigure}
    \caption{
      Here are eight instances of distorted images, produced using the method suggested in~\citep{hendrycksBenchmarkingNeuralNetwork2019}.
    }
    \label{fig:distorted_images_large}
  \end{center}
  \vskip -0.2in
\end{figure*}

\begin{table*}
  \caption{Ablations of the test data distribution on ViT-B/16 (for all methods, $\lambda=0.3$).}
  \label{table:abalation_data_distribution_vit_b_16}
  \begin{center}
    \fontsize{8}{9}\selectfont
    \begin{sc}
      \begin{tabular}{l|ccccc|ccccc}
        \toprule
        \textbf{Method}          & \textbf{Cars}                                           & \textbf{EuroSAT}                                          & \textbf{RESISC45} & \textbf{GTSRB} & \textbf{Avg.} & \textbf{Cars} & \textbf{EuroSAT} & \textbf{RESISC45} & \textbf{GTSRB} & \textbf{Avg.} \\
        \midrule
                                 & \multicolumn{5}{c|}{{Clean Test Set}}                   & \multicolumn{5}{c}{{Corrupted Test Set (Motion Blur)}}                                                                                                                                                 \\
        Task Arithmetic          & 75.3                                                    & 96.3                                                      & 85.3              & 80.5           & 84.3          & 73.5          & 70.9             & 83.9              & 72.2           & 75.1          \\
        Ties-Merging             & 74.8                                                    & 93.4                                                      & 84.0              & 65.8           & 79.5          & 73.1          & 65.5             & 82.3              & 57.4           & 69.6          \\
        AdaMerging               & {83.4}                                                  & {97.2}                                                    & {88.6}            & {97.5}         & {91.7}        & {81.3}        & 75.9             & {87.4}            & {95.6}         & {85.0}        \\
        \textbf{WEMoE (0-layer)} & 78.6                                                    & 95.3                                                      & 86.3              & 95.9           & 89.0          & 75.8          & 73.3             & 84.8              & 91.1           & 81.2          \\
        \textbf{WEMoE (2-layer)} & \textbf{87.3}                                           & \textbf{99.3}                                             & \textbf{96.2}     & \textbf{99.3}  & \textbf{95.5} & \textbf{86.3} & \textbf{76.8}    & \textbf{95.2}     & \textbf{98.2}  & \textbf{89.1} \\
        \midrule
                                 & \multicolumn{5}{c|}{Corrupted Test Set (Impluse Noise)} & \multicolumn{5}{c}{Corrupted Test Set (Gaussian Noise)}                                                                                                                                                \\
        Task Arithmetic          & 70.4                                                    & \textbf{59.5}                                             & 75.2              & 54.0           & 64.8          & 72.2          & \textbf{60.8}    & 78.5              & 51.0           & 65.6          \\
        Ties-Merging             & 70.5                                                    & 46.2                                                      & 73.0              & 42.0           & 57.9          & 72.8          & 47.6             & 77.0              & 42.2           & 59.9          \\
        AdaMerging               & 77.6                                                    & 42.1                                                      & 81.9              & 90.2           & \textbf{73.0} & 79.1          & 58.9             & 81.2              & \textbf{74.5}  & \textbf{73.4} \\
        \textbf{WEMoE (0-layer)} & 72.7                                                    & 47.0                                                      & 77.2              & 80.5           & 69.4          & 74.1          & 58.3             & 80.1              & 64.7           & 69.3          \\
        \textbf{WEMoE (2-layer)} & \textbf{83.2}                                           & 11.1                                                      & \textbf{92.3}     & \textbf{96.2}  & 70.7          & \textbf{84.8} & 11.7             & \textbf{94.4}     & 73.3           & 66.1          \\
        \midrule
                                 & \multicolumn{5}{c|}{Corrupted Test Set (Pixelate)}      & \multicolumn{5}{c}{Corrupted Test Set (Spatter)}                                                                                                                                                       \\
        Task Arithmetic          & 3.8                                                     & 38.0                                                      & 24.8              & 71.3           & 34.5          & 72.1          & 58.4             & 79.9              & 60.1           & 67.6          \\
        Ties-Merging             & \textbf{4.9}                                            & 36.3                                                      & 21.4              & 57.6           & 30.1          & 72.1          & 50.7             & 77.8              & 46.9           & 61.9          \\
        AdaMerging               & 4.1                                                     & \textbf{46.4}                                             & {23.6}            & {91.3}         & \textbf{41.3} & {79.3}        & \textbf{60.9}    & {85.8}            & {93.7}         & \textbf{80.0} \\
        \textbf{WEMoE (0-layer)} & 3.1                                                     & 42.4                                                      & \textbf{26.1}     & 84.3           & 39.0          & 73.7          & 57.              & 82.0              & 87.1           & 74.9          \\
        \textbf{WEMoE (2-layer)} & 0.5                                                     & 20.6                                                      & 1.9               & \textbf{97.3}  & 30.1          & \textbf{84.0} & 11.9             & \textbf{93.8}     & \textbf{97.5}  & 71.8          \\
        \midrule
                                 & \multicolumn{5}{c|}{Corrupted Test Set (Contrast)}      & \multicolumn{5}{c}{Corrupted Test Set (JPEG Compression)}                                                                                                                                              \\
        Task Arithmetic          & 73.4                                                    & 62.5                                                      & 81.3              & 76.9           & 73.5          & 75.1          & 73.1             & 84.8              & 64.7           & 74.4          \\
        Ties-Merging             & 73.4                                                    & 58.0                                                      & 80.0              & 63.1           & 68.6          & 74.8          & 66.9             & 83.8              & 54.1           & 69.9          \\
        AdaMerging               & {81.4}                                                  & \textbf{68.1}                                             & {85.8}            & 96.8           & 83.0          & {81.9}        & 76.0             & 87.3              & {91.0}         & {84.1}        \\
        \textbf{WEMoE (0-layer)} & 76.2                                                    & 65.6                                                      & 82.2              & 93.6           & 79.4          & 77.3          & 74.2             & 86.1              & 84.1           & 80.4          \\
        \textbf{WEMoE (2-layer)} & \textbf{86.0}                                           & 67.8                                                      & \textbf{95.1}     & \textbf{98.9}  & \textbf{87.0} & \textbf{86.9} & \textbf{82.0}    & \textbf{96.2}     & \textbf{95.9}  & \textbf{90.2} \\
        \bottomrule
      \end{tabular}
    \end{sc}
  \end{center}
  \vskip -0.1in
\end{table*}

\begin{figure*}
  \vskip 0.2in
  \begin{center}
    \includegraphics[width=\linewidth]{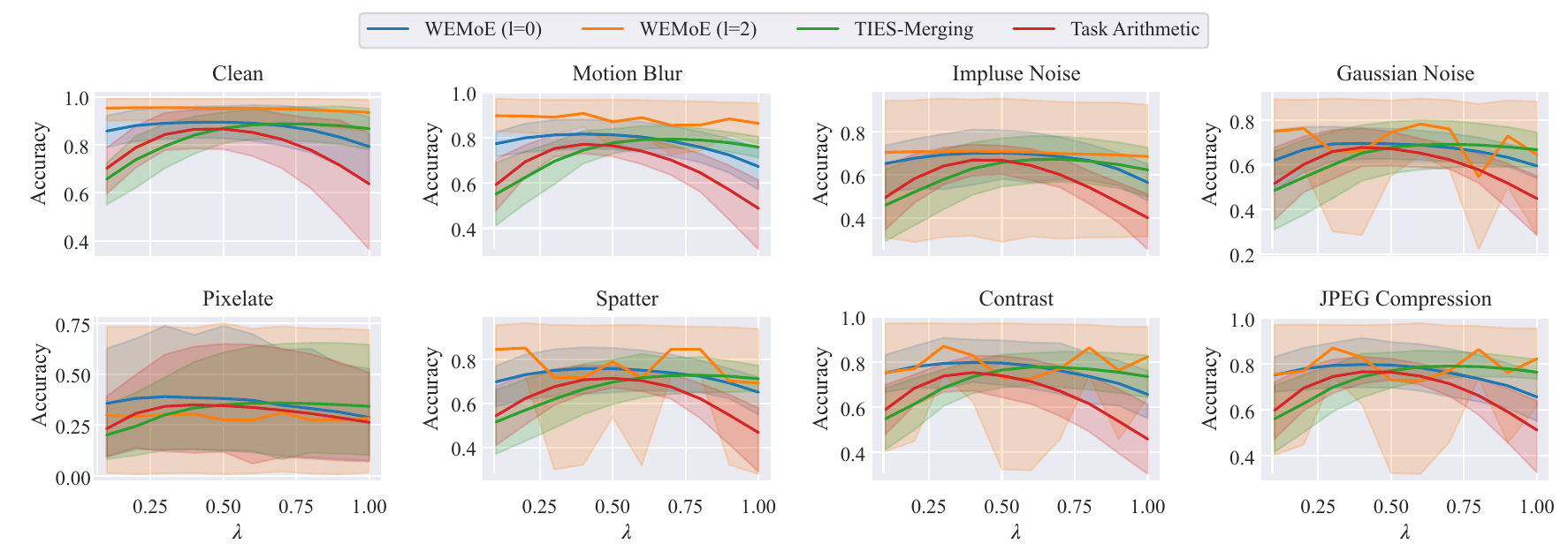}
    \vskip -0.2in
  \end{center}
  \caption{
    The results for the robustness experiment on CLIP-ViT-B/16.
    The x-axis of each plot represents the scaling coefficient $\lambda$ of task vectors, while the y-axis shows the accuracy of the merged model on different merged tasks.
  }
  \label{fig:b16_robustness_lambda}
\end{figure*}

In real-world applications, it is common to encounter out-of-distribution (OOD) data, i.e. the unlabeled test data that we seek to generalize may deviate from the distribution of the training data. 
To evaluate the robustness of our method, we conduct experiments on the clean test dataset and seven corrupted test datasets. 
The corrupted test datasets are generated using the method suggested in~\citep{hendrycksBenchmarkingNeuralNetwork2019}. 
In Figure~\ref{fig:distorted_images_large}, we show eight example images from the corrupted test datasets. 
Following~\citep{yangAdaMergingAdaptiveModel2023}, we conduct experiments on four datasets: Cars~\citep{stallkamp_man_2012}, EuroSAT~\citep{helber2018introducing}, RESISC45~\citep{cheng_remote_2017}, and GTSRB~\citep{stallkamp_man_2012}.

We compare our method with the Task Arithmetic~\citep{ilharcoEditingModelsTask2023}, Ties-Merging~\citep{yadavResolvingInterferenceWhen2023}, and AdaMerging~\citep{yangAdaMergingAdaptiveModel2023}. 
We first set the scaling factor $\lambda=0.3$ and merge the models trained on the clean test dataset.
The results are shown in Tables~\ref{table:abalation_data_distribution_vit_b_32} and~\ref{table:abalation_data_distribution_vit_b_16}. 

The results in Table~\ref{table:abalation_data_distribution_vit_b_32} and Table~\ref{table:abalation_data_distribution_vit_b_16} that among various approaches, WEMoE ($l=2$) consistently demonstrates the highest performance across both the clean test set and the majority of distorted test sets. 
This underscores the effectiveness of our method in handling both clean and distorted data, suggesting its potential for robust multi-task merging.
We also note that in situations where there is a significant degradation in image quality, such as pixelation, WEMoE ($l=2$) may exhibit overfitting to some specific tasks, leading to a decline in performance. In contrast, WEMoE ($l=0$) tends to display greater stability in performance under such conditions. This can be attributed to its lower parameter count, rendering it less susceptible to overfitting.

Furthermore, we note that the performance of WEMoE ($l=0$) closely aligns with that of AdaMerging. It is worth recalling that when $l=0$, WEMoE corresponds to a partial implementation of AdaMerging, with the scaling factor $\lambda$ fixed at $0.3$ for the majority of the model, except for MLP modules.
This observation provides support for the hypothesis outlined in Section~\ref{subsection:revisiting_model_merge} concerning parameter similarity.

Figures~\ref{fig:b32_robustness_lambda} and~\ref{fig:b16_robustness_lambda} show more detailed results of the robustness experiments on CLIP-ViT-B/32 and CLIP-ViT-B/16, respectively. 
The x-axis of each plot represents the scaling coefficient $\lambda$ of task vectors, while the y-axis shows the accuracy of the merged model on the four merged tasks.
The results show that WEMoE ($l=2$) outperforms other methods across all tasks and datasets in the majority of cases. It also exhibits stability across variations in hyperparameter $\lambda$. It achieves nearly optimal performance for all configurations of $\lambda$ when the test dataset follows the same distribution as the training data.
Even when there is a significant disparity between the distributions of test and training data, comparable performance can be achieved with Task Arithmetic and Ties-Merging.

\FloatBarrier

\section{Routing Analysis}
\label{appendix:routing_analysis}

\begin{figure}[hbt]
  \centering
  \includegraphics[width=\linewidth]{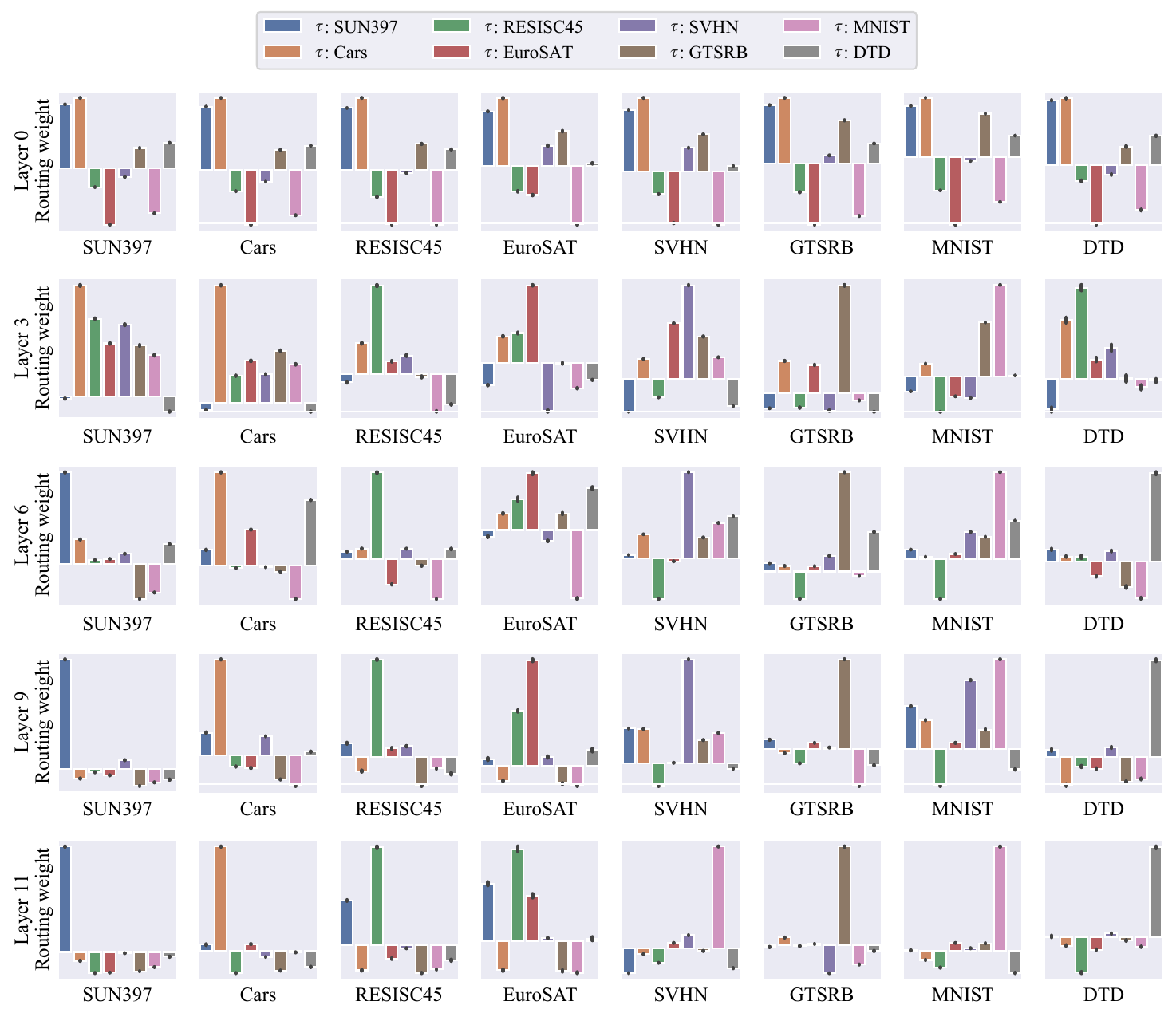}
  \caption{Analysis of expert weighting by the router using the CLIP-ViT-B/32 model at layers 0,3,6,9 and 11.
    This figure presents the routing weights for different layers and tasks in the neural network. Each subplot corresponds to a specific task, and the y-axis represents the routing weights for that task. The x-axis labels indicate the task names.
  }
  \label{fig:router_analysis}
\end{figure}

In this section, we delve into an analysis of the expert weighting as determined by the router. For this analysis, we use the CLIP-ViT-B/32 model that has been merged on eight downstream tasks as a representative example. The results of this analysis are visually represented in Figure~\ref{fig:router_analysis}.
In the figure, we demonstrate how routers at different depths in the network allocate routing weights for inputs from different tasks. This visualization helps in understanding how the routing weights vary across different tasks and layers, providing insights into the network's decision-making process.

Upon examining the router's behavior, we notice a distinct pattern.
For shallow-layer routers, there is no clear correlation between routing weight allocation and the source task of the input sample. Input samples from different tasks exhibit similar routing weights.
In contrast, deep-layer routers show a pronounced correlation between weight allocation and the source task of the input sample. In these cases, the router tends to favor the task vector corresponding to the input sample's source task by assigning it a higher weight. This observation suggests that deeper-layer routers possess the capability to identify the expert likely to perform better based on input features.
As a result, they assign a higher weight to this expert, reflecting increased confidence in its performance.
This behavior of the router is in line with our expectations and intuition.
This ability to discern and assign weights effectively is a key factor in the overall performance of the merged model.

This also indicates that shallow-level features are insufficient to distinguish between different tasks; instead, these features predominantly contain shared information.
In contrast, deep-level features contain more task-specific information. This provides us with insights, suggesting that we can further narrow down the scope of information separation and integration.
For instance, we might consider applying methods like Task Arithmetic to merge parameters of shallow-level MLPs as well, while constructing MoE for information separation exclusively in deep-level MLPs.
This approach can further reduce the inference cost overhead by further reducing additional parameters introduced by MoE while maintaining the performance of the merged model.

\begin{figure}[ht]
  \centering
  \includegraphics[width=0.8\linewidth]{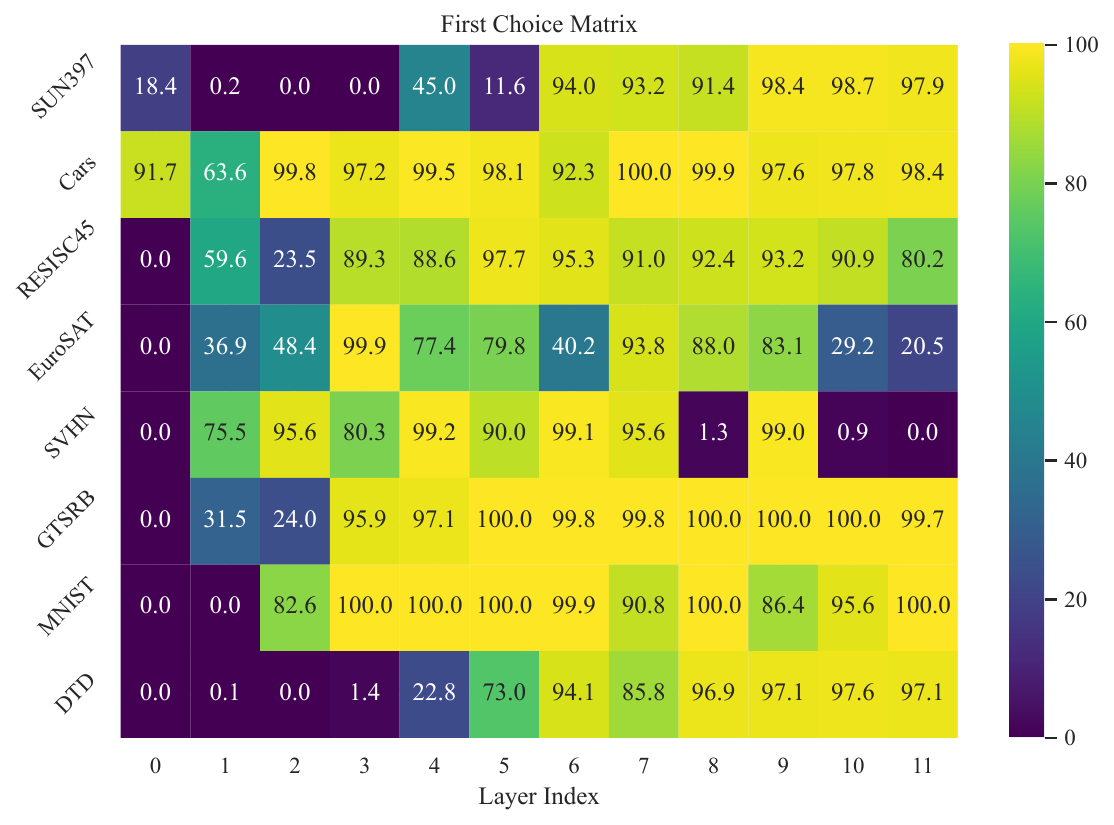}
  \caption{
    This heatmap presents the first choice matrix of the CLIP-ViT-B/32 model.
    Each row corresponds to a specific task, and the x-axis labels indicate the layer index.
    Each entry in the matrix represents the percentage of samples for which the router assigns the highest weight to the corresponding task.
  }
  \label{fig:first_choice_matrix}
\end{figure}

Figure~\ref{fig:first_choice_matrix} presents the first choice matrix of the CLIP-ViT-B/32 model. This matrix indicates the percentage of samples for which the router assigns the highest weight to the corresponding task.
At the first layer, the router assigns almost all samples to Cars. 
As the layer index increases, the router gradually assigns more samples to the correct task.
From the 6th layer onwards, the router assigns the highest weight to the correct task for the majority of samples.
This is consistent with our previous observation from Figure~\ref{fig:router_analysis}, where we notice that the router at the 6th layer is the first to exhibit a clear correlation between routing weight allocation and the source task of the input sample.

\textbf{Why not perform a qualitative analysis of features from various experts?} 
Such an analysis is often beneficial for a comprehensive understanding. 
However, visualizing features from different experts in our study is not straightforward. 
This difficulty arises because our proposed Mixture of Experts (MoE) design differs from research that analyzes features, such as \citet{yeTaskExpertDynamicallyAssembling2023}. 
Our method generates input-conditioned weights and infers through the MLP modules only once. In contrast, \citet{yeTaskExpertDynamicallyAssembling2023} pass the input through each expert. 
Instead, in this Appendix section, we offer a routing analysis that provides insights into the routing mechanism of Weight-Ensembling MoE (WEMoE).

\stopcontents[sections]  %

\end{document}